\documentclass[preprint,12pt]{elsarticle}
\usepackage{amssymb}
\usepackage{xcolor}
\usepackage{amsmath}
\usepackage{booktabs} % 导言区引入一次即可
\usepackage{multirow}

\usepackage{adjustbox}

\usepackage{subcaption}

\journal{Biomedical Signal Processing and Control}

\begin{document}

\begin{frontmatter}



\title{MSAIC-Net: A Multi-Scale Attention and Imbalance-Aware Contrastive Network for ECG-Based Myocardial Substrate Abnormality Detection}

% \affiliation[label1]{organization={},

% \affiliation[label2]{organization={},

% \author{Canyu Lei} %% Author name
% \author{Fenglin Zhang}
% \author{Derek Bivona}
% \author{Cris Singulane}
% \author{Jonathan Pan}
% \author{Amit R. Patel}
% \author{Jianxin Xie}
% \affiliation{organization={University of Virginia},%Department and Organization
%             city={Charlottesville},
%             postcode={22903}, 
%             state={Virginia},
%             country={USA}}

% \affiliation{organization={University of Virginia},%Department and Organization
%             city={Charlottesville},
%             postcode={22903}, 
%             state={Virginia},
%             country={USA}}

\author[label1]{Canyu Lei}
\author[label2]{Fenglin Zhang}
\author[label3]{Derek Bivona}
\author[label3]{Cristiane Singulane}
\author[label3]{Jonathan Pan}
\author[label3]{Kenneth Bilchick}
\author[label3]{Amit R. Patel}
\author[label1]{Jianxin Xie\corref{cor1}}

\cortext[cor1]{Corresponding author.}

\affiliation[label1]{
    organization={School of Data Science, University of Virginia},
    city={Charlottesville},
    postcode={22903},
    state={VA},
    country={USA}
}

\affiliation[label2]{
    organization={Department of Computer Science, University of Virginia},
    city={Charlottesville},
    postcode={22903},
    state={VA},
    country={USA}
}

\affiliation[label3]{
    organization={School of Medicine, University of Virginia},
    city={Charlottesville},
    postcode={22903},
    state={VA},
    country={USA}
}

%% Abstract
\begin{abstract}
Myocardial substrate abnormalities, such as myocardial scar and myocardial infarction (MI), are associated with adverse cardiovascular outcomes. Electrocardiography (ECG) provides a low-cost and widely available tool for detecting these abnormalities, but ECG-based detection remains challenging due to heterogeneous lead-dependent manifestations, high-dimensional multi-lead signals, class imbalance, and the limited interpretability of deep learning models.
We propose a multi-scale attention-enhanced convolutional network (MSAIC-Net) for ECG-based myocardial substrate abnormality detection. MSAIC-Net employs parallel atrous convolutional branches to extract ECG features across multiple temporal receptive fields. %, enabling the model to capture both local and longer-range temporal patterns. 
Channel attention is then used to adaptively reweight informative lead-wise and feature-channel representations. To address class imbalance and improve feature separability, we introduce a novel imbalance-aware supervised contrastive learning strategy that encourages samples from the same class to form compact representations while increasing separation between abnormal and normal samples. Lead-wise permutation importance is further incorporated to quantify the contribution of each ECG lead and improve model interpretability.
The proposed method was evaluated on two complementary datasets: a low-data institutional cohort from the University of Virginia (UVA) Health System for myocardial scar classification and the large-scale public PTB-XL dataset from PhysioNet for MI identification. Experimental results show that MSAIC-Net outperforms baseline models, with particularly pronounced improvements in the low-data UVA cohort. Overall, the proposed framework provides an effective and interpretable approach for ECG-based detection of myocardial substrate abnormalities.

\end{abstract}

\begin{graphicalabstract}
\includegraphics[width=\textwidth]{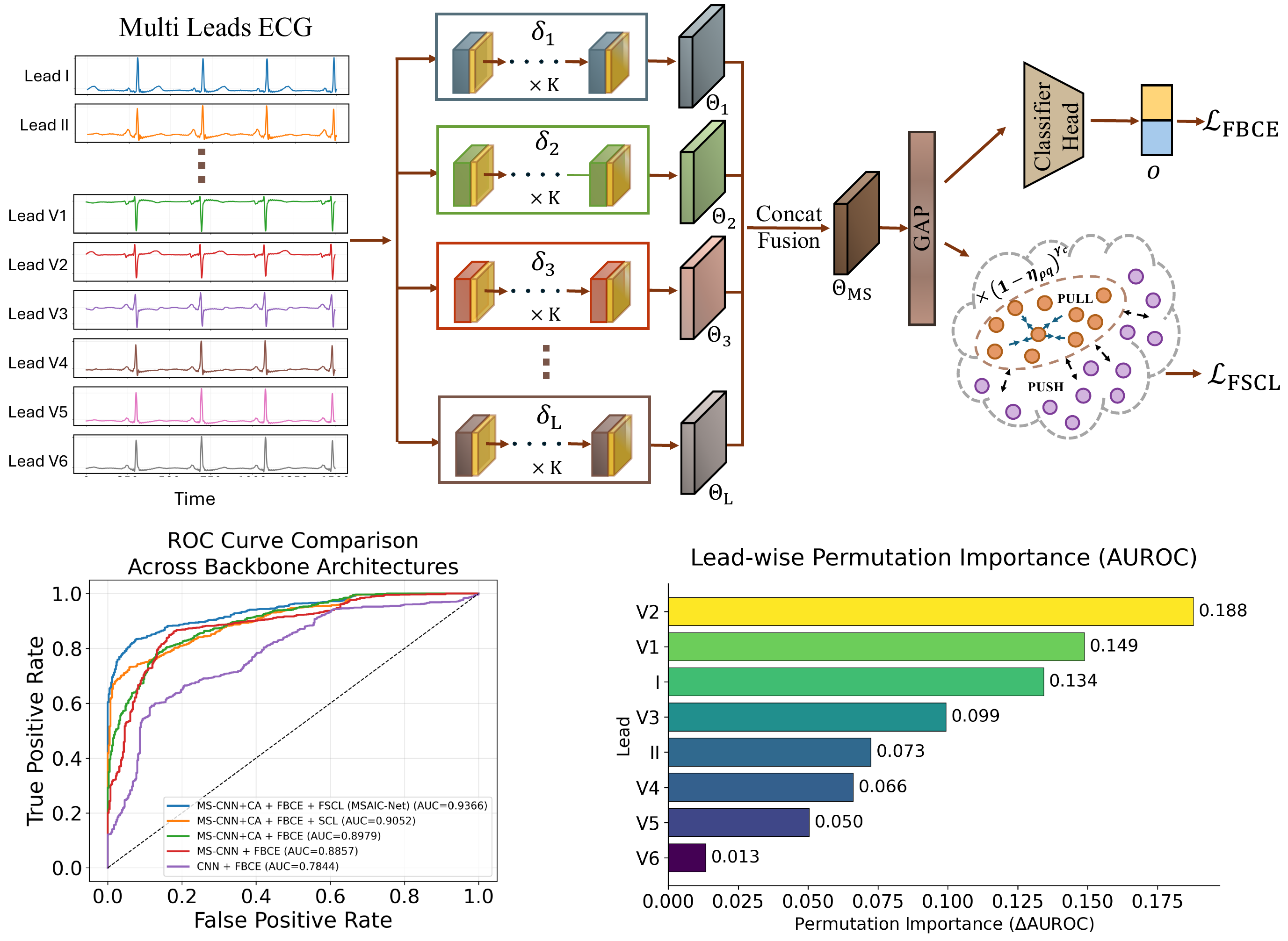}
\end{graphicalabstract}

\begin{highlights}
    \item MSAIC-Net enables ECG-based substrate abnormality detection.
    \item Multi-scale attention improves ECG feature representation.
    \item Focal contrastive learning enhances class separability.
    \item Lead-wise permutation importance supports interpretability.
\end{highlights}

%% Keywords

\end{frontmatter}



\section{Introduction}
\label{intro}

Among cardiac abnormalities, myocardial substrate abnormalities, including myocardial scar, fibrosis, and myocardial infarction (MI)-related tissue injury, represent structural remodeling of the myocardium caused by ischemic, inflammatory, or non-ischemic cardiomyopathic processes~\cite{richardson2015physiological}. These abnormalities alter both the mechanical and electrical properties of cardiac tissue and may create an arrhythmogenic substrate associated with adverse outcomes, including ventricular arrhythmias, heart failure progression, and sudden cardiac death~\cite{strauss2008ecg,acosta2018scar}. Given their substantial clinical and public health burden, early detection of myocardial substrate abnormalities is of critical importance for preventing adverse clinical outcomes~\cite{de2010early,baghdadi2023advanced,celermajer2012cardiovascular}.
Myocardial substrate abnormalities are typically evaluated using cardiac imaging. Among available modalities, late gadolinium enhancement cardiac magnetic resonance (LGE-CMR) is widely used for non-invasive myocardial tissue characterization and is considered a reference standard for assessing focal scar, replacement fibrosis, and infarct-related myocardial injury~\cite{holtackers2022late,meier2024myocardial}. LGE-CMR can characterize the presence, extent, and spatial distribution of abnormal myocardial tissue, which is important for diagnosis and treatment planning. Nevertheless, routine LGE-CMR-based screening is constrained by high cost, scanner availability, and contrast-agent requirements, leading to limited applicability~\cite{holtackers2022late,scholtz2014current}. %In contrast, 12-lead ECG is inexpensive, rapid, and routinely acquired in clinical care. Developing ECG-based artificial intelligence models for myocardial substrate abnormality detection may therefore provide a scalable strategy to identify high-risk patients and guide subsequent imaging-based evaluation.

Because myocardial substrate abnormalities can disturb depolarization and repolarization, their effects may be reflected in surface electrocardiograms (ECGs), such as abnormal QRS morphology, pathological Q waves, ST--T changes, or other lead-dependent electrical patterns~\cite{strauss2008ecg,gordeeva2022electrocardiographic}. Therefore, ECG provides a low-cost and widely available modality with potential value for screening and risk assessment of myocardial substrate abnormalities. 
Traditional ECG-based detection of myocardial infarction or scar-related abnormalities has largely relied on visually defined or handcrafted ECG features~\cite{michael2007electrocardiographic,strauss2009qrs}. 
Although these features are clinically meaningful, they may not fully capture complex substrate-related ECG manifestations, which can be subtle, spatially heterogeneous, and dependent on the location and extent of the abnormal substrate~\cite{xiong2022deep,miranda2018new}.

With the advancement of machine learning and deep learning, data-driven approaches have been increasingly explored to automatically detect myocardial infarction and related abnormalities from ECG signals. Compared with handcrafted feature-based methods, deep learning models can directly learn discriminative representations from raw or minimally processed ECG waveforms. For example, convolutional neural network (CNN)-based models have been widely used to extract local temporal patterns from ECG signals and have shown promising performance in automated ECG diagnosis and MI detection~\cite{xiong2022deep,ribeiro2020automatic}. 

%With the continuous advancement of deep learning techniques, data-driven approaches have been increasingly explored for ECG-based myocardial scar detection~\cite{xiong2022deep,sumalatha2024deep}.

Despite these advances, several important challenges remain for ECG-based myocardial substrate abnormality detection. Many existing CNN-based ECG models rely on fixed-size convolutional kernels or fixed receptive fields, which may limit their ability to capture scar-related ECG patterns occurring at different temporal scales~\cite{jian2021detection,feng2025ms}. For example, substrate abnormalities may induce localized changes in the QRS complex, fragmented conduction patterns, ST-segment deviations, or broader repolarization abnormalities~\cite{gordeeva2022electrocardiographic,michael2007electrocardiographic}. Although 12-lead ECGs provide complementary views of cardiac electrical activity, standard deep learning models may not fully exploit lead-dependent information or adaptively emphasize the most informative leads and feature channels~\cite{cao2022detection}. %Third, clinical scar datasets are often limited in size and imbalanced, making models vulnerable to majority-class bias and reducing sensitivity to subtle scar-positive cases. Finally, many deep learning models provide limited interpretability, which restricts their clinical utility for understanding how different ECG leads contribute to scar-related predictions.

Additionally, class imbalance is a key challenge in ECG-based myocardial abnormality detection, where the number of samples varies significantly across classes~\cite{lu2018feature}. Existing approaches typically address class imbalance through data-level resampling strategies, such as oversampling the minority class, undersampling the majority class, or generating additional minority-class samples using synthetic data generation models. Other approaches built specialized model architectures~\cite{xie2024automated, pandey2019automatic,rath2021heart}. However, resampling-based methods may distort the original data distribution, synthetic data generation methods may produce physiologically unrealistic samples, and specialized model designs may introduce additional computational complexity~\cite{choi2023performance}. Therefore, learning robust and discriminative representations under imbalanced data remains essential for reliable ECG-based myocardial abnormality detection.

In addition to class imbalance, model interpretability remains an important challenge. Although 12-lead ECGs provide spatially distributed views of cardiac electrical activity, deep learning models operate as black boxes and offer limited insight into how individual leads contribute to the final prediction~\cite{van2021relation,zhang2021interpretable,hong2019mina}. Interpretability is particularly important for myocardial substrate abnormality detection, because lead-level patterns may reflect regional differences in scar or infarct-related electrical changes~\cite{jahmunah2022explainable}. Prior ECG studies %have explored lead contribution in ECG deep learning using approaches such as multi-lead attention~\cite{fu2020hybrid} and SENet-based lead weighting~\cite{cao2022detection}. These studies 
suggest that different ECG leads contribute unequally to model predictions and that lead-level attribution may provide clinically relevant information about regional cardiac electrical abnormalities 
~\cite{fu2020hybrid,cao2022detection}. %However, lead-wise explainability remains underexplored for substrate abnormality detection, particularly in models that quantify the contribution of each lead in post-training phase rather than relying only on internal attention weights. In this study, we perform post-training lead-wise explainability analysis using permutation importance to quantify the contribution of each ECG lead to the model’s prediction. Compared with gradient-based methods such as Grad-CAM or saliency maps, which often highlight local waveform segments and can be sensitive to model architecture and gradient behavior, permutation importance provides a model-agnostic and performance-based estimate of lead contribution by measuring the change in prediction performance after perturbing each lead. 
However, lead-wise explainability remains underexplored for myocardial substrate abnormality detection, particularly in models that quantify the contribution of each lead after training rather than relying only on internal attention weights~\cite{wagner2024explaining}. Although attention mechanisms can indicate how the model adaptively weights feature channels during representation learning, attention scores do not necessarily provide a performance-based estimate of how much each ECG lead contributes to the final prediction~\cite{jain2019attention,wiegreffe2019attention}. Therefore, post-training interpretability is needed to directly assess the functional importance of each lead by evaluating how model predictions change when lead-specific information is perturbed~\cite{fisher2019all}.

%In this study, we perform post-training lead-wise explainability analysis using permutation importance to quantify the contribution of each ECG lead to the model's prediction. Compared with gradient-based methods such as Grad-CAM or saliency maps, which often highlight local waveform segments and can be sensitive to model architecture and gradient behavior, permutation importance provides a model-agnostic and performance-based estimate of lead contribution by measuring the change in prediction after perturbing each lead.
%Permutation importance has been used in prior ECG-based machine learning studies, particularly for interpreting handcrafted ECG features or conventional classifiers. However, its use for post-training lead-wise explainability in deep learning remains relatively underexplored. In this study, we use lead-level permutation importance to quantify how disrupting each ECG lead affects scar classification performance.

To address the aforementioned challenges, we propose a multi-scale attention-enhanced convolutional network (MSAIC-Net) for myocardial abnormality detection from class-imbalanced multi-lead ECG signals. The proposed method integrates adaptive feature weighting and contrastive regularization to effectively extract discriminative features associated with myocardial disease from high-dimensional multi-lead signals. We further perform post-training explainability analysis using permutation importance to quantify lead-level contributions and identify the ECG leads most influential to the final classification decision.
The main contributions of this work are summarized as follows:
\begin{itemize}
    \item We leverage multi-scale convolutional operations to capture both local and global temporal patterns, while incorporating attention mechanisms to adaptively emphasize informative leads and channels, thereby enhancing feature representation.

    \item To mitigate training bias caused by class imbalance, we employ focal binary cross-entropy loss and further propose a focal-weighted supervised contrastive learning strategy. The proposed focal contrastive objective assigns larger weights to hard positive pairs during representation learning.This encourages discriminative feature embeddings with improved intra-class compactness and inter-class separability, thereby enhancing myocardial abnormality detection performance under imbalanced data.

    % \item To enhance interpretability, we perform post-training lead-wise explainability analysis using permutation importance to quantify the contribution of each ECG lead to the model’s final classification decision. Because different ECG leads reflect electrical activity from different cardiac regions, the identified lead-importance patterns provide cohort-level insight into which ECG leads are most informative for myocardial substrate abnormality classification and may offer indirect clues about the regional myocardial abnormalities represented in the study cohort.
    \item To enhance interpretability, we perform post-training lead-wise explainability analysis using permutation importance to quantify the contribution of each ECG lead to the final classification decision. The identified lead-importance patterns provide cohort-level insight into which ECG leads are most informative for myocardial substrate abnormality detection and may offer indirect clues about regional myocardial abnormalities represented in the study cohort.
\end{itemize}

\section{Related Work}

Early ECG analysis predominantly relied on manually engineered or semi-automatically extracted features guided by domain expertise~\cite{de2004automatic,clifford2006advanced,hu1993applications}. 
These approaches typically extract handcrafted features based on clinical knowledge, such as QRS complex morphology, ST-segment deviations, T-wave abnormalities, and QT intervals, often aided by signal processing techniques such as time-frequency analysis and dimensionality reduction~\cite{mark1988bih,yu2007electrocardiogram,pan2007real,saxena2002feature,wacker2013time}. 
These handcrafted features are often combined with traditional machine learning models such as support vector machines and decision trees for classification tasks~\cite{osowski2004support,llamedo2010heartbeat,sarkaleh2012classification}. Since these features are closely related to established clinical indicators, they provide a certain degree of interpretability.
However, such methods have notable limitations, as handcrafted feature design depends heavily on expert knowledge, and feature extraction is often labor-intensive and time-consuming \cite{montenegro2022human,ding2024deep}. 

% CNN
With the rapid development of deep learning, ECG analysis has increasingly shifted from handcrafted feature extraction to representation learning directly from ECG signals. These models can automatically extract latent features and have demonstrated superior performance in various ECG classification tasks. For instance, convolutional neural networks (CNNs) are among the most widely used architectures in ECG analysis, as they can capture local morphological patterns and temporal dependencies from time series signals. 
Among them, 1D-CNNs have achieved strong performance in ECG classification tasks by automatically learning discriminative temporal and morphological representations directly from raw ECG signals.
For example, Baloglu et al. developed an end-to-end convolutional neural network (CNN) model for automatic myocardial infarction detection from standard 12-lead ECG signals, demonstrating high diagnostic performance with accuracy and sensitivity exceeding 99\%~\cite{baloglu2019classification}. 
Ozaltin et al. proposed a hybrid CNN--SVM framework for ECG classification, where continuous wavelet transform (CWT) transformed one-dimensional ECG signals into scalogram images, and CNN-extracted features were subsequently classified using an SVM~\cite{ozaltin2023novel}. Following this CWT-based representation strategy, Xie et al. developed a multi-branch ResNet architecture to capture complementary time-frequency ECG patterns for automated atrial fibrillation detection~\cite{xie2024automated}. Khan et al.~\cite{khan2023ecg} proposed a one-dimensional ResNet-based framework for ECG classification. Through residual learning and hierarchical feature extraction, the model achieved high diagnostic accuracy on the MIT-BIH Arrhythmia Database, demonstrating the effectiveness of deep convolutional architectures for ECG analysis. However, conventional CNNs typically rely on fixed-size convolutional kernels, which are inherently biased toward local pattern extraction and may require deeper architectures or additional mechanisms to capture long-range temporal and inter-lead dependencies.

% RNN
Since ECG signals are inherently temporal sequences, recurrent neural networks (RNNs) and their variants, such as long short-term memory (LSTM) and gated recurrent units (GRU), have also been widely applied to ECG time-series modeling. 
For example, Singh et al. applied RNNs for ECG arrhythmia classification, demonstrating the capability of RNN-based models in distinguishing normal and abnormal heartbeats~\cite{singh2018classification}. Xiong et al. proposed a convolutional recurrent neural network (CRNN) for ECG classification, where convolutional layers extract local features and recurrent layers model temporal dependencies for arrhythmia detection~\cite{xiong2018ecg}. 
Alamatsaz et al.~\cite{alamatsaz2024lightweight} developed a hybrid CNN-LSTM framework for ECG classification, demonstrating the effectiveness of combining convolutional feature extraction with temporal sequence modeling for cardiac abnormality detection.
Zhang et al.~\cite{zhang2019new} developed an LSTM-based approach for myocardial infarction detection, showing that recurrent neural networks can effectively capture temporal patterns in ECG signals for cardiac abnormality classification.
However, RNN-based models generally suffer from low training efficiency and are prone to gradient-related issues when handling long ECG sequences, which limits their ability to effectively capture long-range dependencies~\cite{satheeswaran2024deep}.

%\paragraph{attention}
In recent years, attention mechanisms and Transformer architectures have been increasingly introduced into ECG analysis. 
For example, Satheeswaran et al. proposed an attention-enhanced LSTM model for ECG classification, where the attention mechanism enables the model to focus on important temporal segments, improving both performance and interpretability~\cite{satheeswaran2024deep}. 
Mousavi et al. proposed an interpretable hierarchical attention network for atrial fibrillation detection, where multi-level attention mechanisms were used to identify the most informative temporal patterns in ECG signals~\cite{mousavi2020han}. 
Hu et al. proposed a Transformer-based model for arrhythmia detection from continuous ECG signals, leveraging self-attention to capture inter-heartbeat dependencies without requiring explicit segmentation~\cite{hu2022transformer}. 
Ikram et al.~\cite{ikram2025transformer} developed a Transformer-based model for ECG arrhythmia classification. The proposed framework demonstrated good classification performance on the MIT-BIH dataset across multiple arrhythmia categories.
However, Transformer-based models typically require large-scale data for stable training and generalization~\cite{takahashi2024comparison}. In medical settings with limited data and costly annotations, complex Transformers may lead to overfitting and high computational cost~\cite{cheng2023msw}. Therefore, lightweight and task-specific architectures are more suitable when labeled clinical ECG datasets are limited in scale.

% Imbalance Data
In addition to model architecture, class imbalance is a common issue in ECG classification tasks, where abnormal samples are typically fewer than normal samples. This imbalance can bias models toward the normal class, resulting in high accuracy but reduced sensitivity to disease cases and an increased risk of false-negative predictions. Existing studies have addressed this issue through data-level resampling, model-structure design, or loss-function modification. For example, Pandey et al. used Synthetic Minority Over-sampling Technique (SMOTE)-based oversampling to augment minority ECG classes~\cite{pandey2019automatic}. 
Rath et al. employed generative adversarial network (GAN)-based augmentation to generate synthetic ECG samples~\cite{rath2021heart}. However, these data-level methods may distort the original data distribution, introduce synthetic artifacts, and increase training cost due to the expanded training set. Structure-level methods have also been explored. For instance, Xie et al. proposed a multi-branch ResNet in which each branch was trained on a balanced subset formed by combining minority-class samples with an equal number of randomly selected majority-class samples~\cite{xie2024automated}. Although effective, such multi-branch designs may require extensive branch-wise training. Loss-level methods, such as focal binary cross-entropy, can reduce majority-class dominance by emphasizing hard samples, but their improvement may be limited if the learned representations remain poorly separated~\cite{lin2025multitask,ammar2025ecg}. Therefore, more effective representation learning strategies are needed to improve class discrimination under imbalanced ECG data.

%\paragraph{Explainability}
Explainability is particularly important for ECG-based detection of myocardial substrate abnormalities, because these abnormalities can alter regional cardiac electrical activity. Existing ECG explainability studies have used several strategies to interpret model decisions. Fu et al. used a multi-lead attention mechanism to assign different weights to 12-lead ECG signals for MI detection and localization~\cite{fu2020hybrid}. Cao et al. incorporated SENet-based attention and Grad-CAM visualization to calculate and visualize lead-level weighting in a multi-scale ResNet framework for MI detection and localization~\cite{cao2022detection}. Jahmunah et al. applied Grad-CAM to CNN- and DenseNet-based models to highlight discriminative ECG regions associated with MI classification~\cite{jahmunah2022explainable}. %More broadly, recent ECG-XAI studies have adopted saliency maps, Grad-CAM, SHAP, and LIME to visualize important signal regions or feature contributions in deep ECG models~\cite{manimaran2025explainable}. 
While these explainability techniques provide useful insights into model behavior, they have several limitations for lead-wise interpretation. Attention-based weights are internal model parameters learned during training and may not directly represent the functional importance of each ECG lead to the final prediction~\cite{jain2019attention,wiegreffe2019attention}. Similarly, Grad-CAM and saliency-based methods are often designed to highlight local waveform regions that influence a prediction, but they can be sensitive to model architecture, gradient behavior, and the selected network layer~\cite{selvaraju2017grad,wagner2024explaining}. Therefore, these methods may not provide a direct performance-based estimate of how much each lead contributes to the model decision.
%While these explainability techniques can provide useful explanations, they often focus on local waveform regions, model-internal activations, or feature-level attribution. 
%To address this limitation, we introduce lead-wise permutation importance as a post-training explainability analysis. 
%In contrast, permutation importance directly evaluates lead-level contribution by measuring the change in classification performance after perturbing each ECG lead~\cite{fisher2019all,hicks2021explaining}. %Therefore, it is particularly suitable for identifying which leads most strongly influence the final classification decision. 
In contrast, permutation importance provides a post-training and model-agnostic strategy for quantifying lead-wise contribution~\cite{fisher2019all,hicks2021explaining}. Rather than interpreting internal attention weights or gradients, it directly evaluates the functional importance of each ECG lead by perturbing that lead and measuring the resulting change in the trained model's prediction.

\section{Methods}

\subsection{Problem Statement}

ECG recordings reflect the electrical activity of the heart and can capture functional abnormalities associated with structural changes such as myocardial scarring and MI.
In this work, we formulate myocardial substrate abnormality detection as a binary classification problem based on multi-lead ECG time-series data. Specifically, let $\mathcal{D} = {(X_{\rho}, y_{\rho})}_{\rho=1}^{N}$ denote the labeled ECG dataset, where $N$ is the number of ECG recordings. For the $\rho$-th sample, $X_{\rho} \in \mathbb{R}^{D \times T}$ represents a multi-lead ECG signal with $D$ leads and $T$ temporal instances. $y_{\rho} \in {0,1}$ denotes the corresponding binary label, with $y_{\rho}=0$ indicating normal class and $y_{\rho}=1$ indicating presence of myocardial abnormality.

The learning objective is to learn a mapping function $f_{\theta}$ that predicts labels from multi-lead ECG time-series signals:
\begin{equation}
    f_{\theta}: \mathbb{R}^{D \times T} \rightarrow [0,1], \quad \hat{y}_{\rho} = f_{\theta}(X_{\rho})
\end{equation}
where $\hat{y}_{\rho}$ represents the predicted probability that the sample belongs to the diseased class. The function $f_{\theta}$ is parameterized by the proposed MSAIC-Net which is designed to learn discriminative and clinically meaningful representations for myocardial abnormality detection.

\subsection{Multi-scale attention-enhanced convolutional
network}

\begin{figure*}[htbp]
\centering
\includegraphics[width=1.0\linewidth]{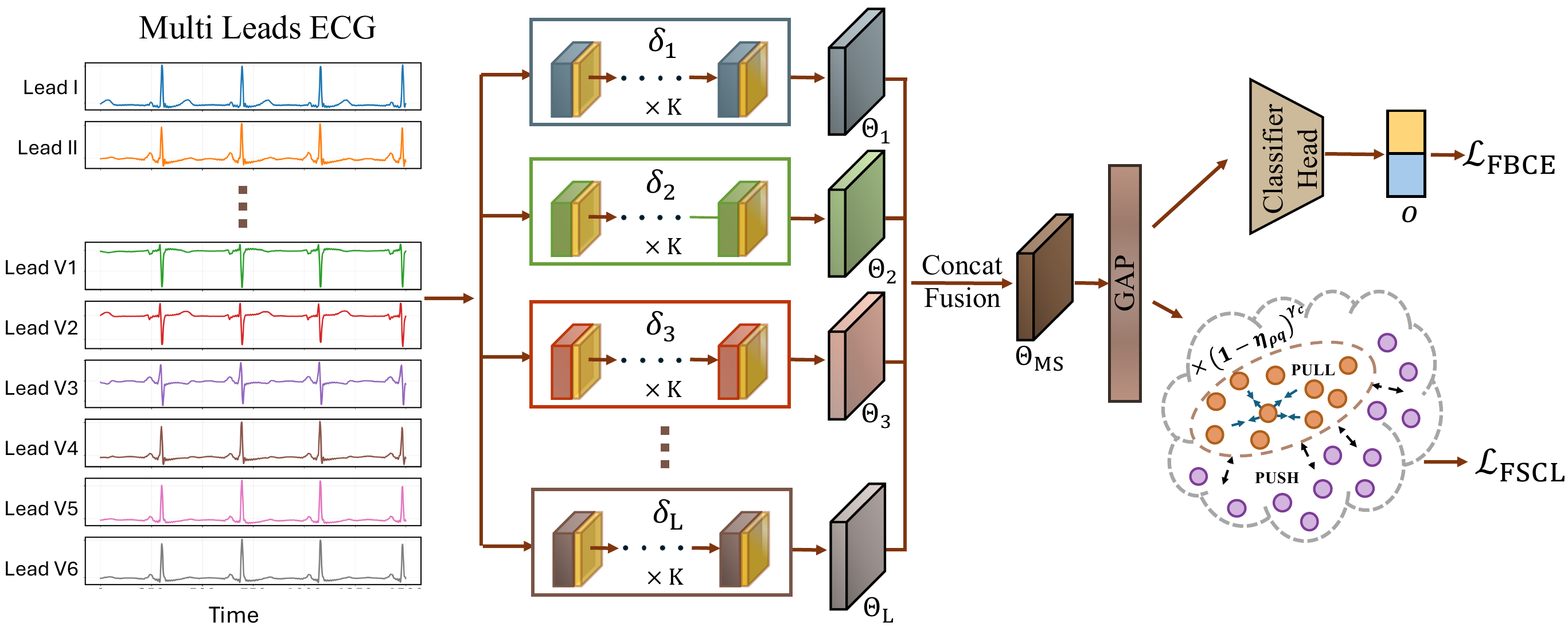}
\caption{Overview of the proposed MSAIC-Net architecture. }
\label{fig:maec}
\end{figure*}

myocardial abnormality detection from multi-lead ECG signals is a clinically important yet challenging task. Scar-related ECG signatures are often subtle and distributed across multiple temporal scales. Some discriminative patterns arise from short-term local waveforms, such as QRS complexes and ST-segment changes, while others are reflected in longer-range temporal dynamics, such as rhythm and conduction patterns. In addition, multi-lead ECG recordings contain physiologically correlated and partially redundant information, requiring the model to capture both intra-lead temporal morphology and cross-lead feature dependencies. However, conventional CNN-based methods are typically constrained by fixed receptive fields and may not sufficiently characterize such multi-scale and channel-dependent ECG patterns, thereby limiting their performance in myocardial abnormality detection.
To address these limitations, we propose MSAIC-Net, a multi-scale attention-enhanced convolutional network for myocardial abnormality detection from multi-lead ECG time-series data, as illustrated in Fig.~\ref{fig:maec}. The core of MSAIC-Net is an integrated multi-branch atrous attention module designed to jointly capture temporal patterns at different receptive fields and adaptively emphasize informative feature channels. 

The MSAIC-Net consists of $L$ parallel branches, where each branch is assigned a unique dilation rate $\delta_l$. Branches with smaller dilation rates focus on fine-grained local waveform morphology, whereas branches with larger dilation rates capture broader temporal dependencies. Within each branch, we propose a novel neural network block, the Attention-Enhanced Atrous Convolution (AEAC) Block, where
each atrous convolution is coupled with channel attention, allowing the model to progressively refine scale-specific ECG representations while recalibrating channel-wise feature importance. Since ECG recordings are one-dimensional time-series signals, 1D convolutional neural networks (1D-CNNs) were used in all experiments.

Specifically, the feature representation extracted by the $l$-th branch is defined as
\begin{equation}
    \Theta_{l} = \mathcal{B}_{l}(X; \delta_l), \quad l = 1,\dots,L,
\end{equation}
where $\mathcal{B}_{l}(\cdot)$ denotes the $l$-th attention-enhanced atrous convolution branch with dilation rate $\delta_l$. Within each branch, $K$ AEAC blocks are stacked to progressively refine scale-specific ECG features, formulated as
\begin{equation}
    \mathcal{B}_l
    =
    \mathcal{A}_l^{(1)}
    \circ
    \cdots
    \circ
    \mathcal{A}_l^{(K-1)}
    \circ
    \mathcal{A}_l^{(K)},
\end{equation}
where $\mathcal{A}_l^{(k)}(\cdot)$ denotes the $k$-th AEAC block in the $l$-th branch. Each AEAC block performs atrous convolution followed by channel-wise attention recalibration:
\begin{equation}
    \mathcal{A}_l^{(k)}(\cdot)
    =
    \mathrm{Att_c}
    \left(U_l^{(k)}
    %\mathrm{Conv}_{\delta_l}(X)
    \right),
\end{equation}
where $U_l^{(k)}=\mathrm{Conv}_{\delta_l}(\cdot)\in\mathbb{R}^{C\times T}$ denotes atrous convolution with dilation rate $\delta_l$, and $\mathrm{Att_c}(\cdot)$ represents the channel attention~\cite{wang2020eca}. $C$ denotes the total number of channels and $T$ denotes the temporal length.

For the channel attention operation $\mathrm{Att_c}$, global average pooling is first applied over the temporal dimension to summarize each channel into a compact descriptor:
\begin{equation}
z_{l,c}^{(k)}
=
\frac{1}{T}
\sum_{t=1}^{T}
U_{l,c}^{(k)}(t),
\qquad
c=1,2,\dots,C.
\end{equation}
The resulting channel descriptor vector is denoted as
\begin{equation}
\boldsymbol{z}_l^{(k)}
=
\left[
z_{l,1}^{(k)}, z_{l,2}^{(k)}, \dots, z_{l,C}^{(k)}
\right]^\top
\in \mathbb{R}^{C}.
\end{equation}
 A one-dimensional convolution is then used to model local cross-channel interactions and generate channel attention weights:
\begin{equation}
w_l^{(k)}
=
\sigma
\left(
\mathrm{Conv1D}(\boldsymbol{z}_l^{(k)})
\right),
\end{equation}
where $\sigma(\cdot)$ denotes the Sigmoid activation. Finally, the attention weights are applied channel-wise to recalibrate the convolution output:
\begin{equation}
\tilde U_{l}^{(k)}
=
\boldsymbol{w}_{l}^{(k)}
\cdot
U_{l}^{(k)},
% \qquad
% c=1,2,\dots,C.
\end{equation}
Through this lightweight cross-channel interaction mechanism, MSAIC-Net enhances discriminative ECG feature learning with minimal additional parameter overhead.

In implementation, each atrous convolution is followed by batch normalization, ReLU activation, and dropout to improve training stability and generalization.
This stacked convolution-attention design allows each branch to learn deeper representations within a fixed temporal scale. Meanwhile, the embedded channel attention mechanism adaptively emphasizes informative channels, which is important because different ECG leads and latent feature channels may contribute unequally to myocardial abnormality detection.

The outputs from all branches are concatenated along the channel dimension and passed through a projection convolution layer to generate the fused multi-scale representation:
\begin{equation}
    \Theta_{\mathrm{MS}}
    =
    \phi
    \left(
    \mathrm{Concat}(\Theta_1, \Theta_2, \dots, \Theta_L)
    \right),
\end{equation}
where $\phi(\cdot)$ denotes the projection convolution operation, which integrates complementary information extracted across different temporal scales and compresses the channel dimension. Global average pooling is then applied along the temporal dimension to aggregate the multi-scale features and obtain a compact channel-wise representation:
\begin{equation}
    \Theta_{\mathrm{f}} = \mathrm{GAP}(\Theta_{\mathrm{MS}}).
\end{equation}
The resulting global representation $\Theta_{\mathrm{f}}$ is fed into the subsequent classification head for myocardial abnormality prediction.

\subsection{Loss Function Design}

To train the proposed model effectively under class imbalance, we adopt a dual-objective loss design consisting of a focal binary cross-entropy loss and a focal contrastive loss. The focal binary cross-entropy (BCE) loss directly supervises the myocardial disease classification task while reducing the dominance of easily classified samples, thereby encouraging the model to focus more on difficult and minority-class cases. However, optimizing the focal BCE loss alone primarily enforces sample-level classification accuracy and does not shape a good decision boundary of the latent feature space. As a result, samples with the same abnormality status may still be dispersed in the embedding space, while samples from different classes may remain insufficiently separated. To address this limitation, we further introduce a focal supervised contrastive loss to enhance intra-class compactness and inter-class separability in the learned representations.

\subsubsection{Focal Binary Cross-entropy Loss}
To enforce consistency between the predicted labels and the ground-truth annotations, we introduce a classification loss as the primary supervision signal for network optimization. This objective directly constrains the model output to align with the target class label. Given the $\rho$-th input ECG sample $X_{\rho}$, the backbone network followed by the classification head produces a prediction logit $\xi_{\rho}$.
% $\zeta_{\rho}\in \mathbb{R}^d$. 
The logit is then transformed into a probability through the sigmoid activation to obtain the final probability
$p_{\rho} = \sigma(\xi_{\rho})$.
Considering that ECG classification tasks often suffer from class imbalance, we adopt focal binary cross-entropy as the classification objective to emphasize hard-to-classify samples~\cite{lin2017focal}. The loss for the $\rho$-th sample is defined as:
\begin{equation}
\mathcal{L}_{\mathrm{FBCE}}^{(\rho)}
=
- \alpha y_{\rho} (1-p_{\rho})^\gamma \log(p_{\rho})
- (1-\alpha)(1-y_{\rho})p_{\rho}^\gamma \log(1-p_{\rho}),
\end{equation}
where $\alpha\in[0,1]$ is the class balancing factor that controls the relative importance of positive and negative samples, and $\gamma>0$ is the focusing parameter that reduces the contribution of easy samples.
The total classification loss is computed as:
\begin{equation}
\mathcal{L}_{\mathrm{FBCE}}
=
\frac{1}{N}\sum_{\rho=1}^{N}\mathcal{L}_{\mathrm{FBCE}}^{(\rho)}.
\end{equation}

This loss directly supervises the prediction logits and serves as the primary classification objective, driving the model predictions toward accurate alignment with the ground-truth labels.

\subsubsection{Supervised Contrastive Strategy}

To further enhance representation learning while improving classification performance, we introduce supervised contrastive learning in addition to the focal BCE objective, enabling the network to learn more discriminative and structured feature representations.

Supervised contrastive learning explicitly models relationships among samples by pulling embeddings of the same class closer together while pushing embeddings of different classes farther apart, thereby improving intra-class compactness and inter-class separability in the feature space and enhancing representation discriminability~\cite{khosla2020supervised}.
However, standard supervised contrastive learning assigns equal weights to all positive pairs without considering their relative learning difficulty. During training, many easy positive pairs already exhibit high similarity yet still dominate gradient contributions, whereas hard positive pairs, those sharing the same label but remaining far apart in feature space, require stronger optimization emphasis but receive insufficient attention under uniform weighting. This limitation becomes more pronounced in ECG signals, where substantial intra-class variation is common.
To address this issue, we propose a Focal Supervised Contrastive Loss, which introduces a focal weighting mechanism to adaptively emphasize hard positive pairs. Specifically, positive pairs with lower matching probabilities are assigned larger loss weights, forcing the model to focus on difficult alignment cases, thereby improving overall representation quality and enhancing robustness to complex ECG patterns.

Formally, consider a mini-batch containing $B$ input ECG samples. After passing through the backbone encoder and projection head, the $\rho$-th sample is mapped into $\zeta_\rho$. Thus, the mini-batch forms a set of $B$ feature embeddings: $\{\zeta_\rho\}_{\rho=1}^B$.
Positive and negative pairs are then constructed based on the ground-truth class labels. For an anchor embedding, embeddings from samples with the same class label are treated as positives, while embeddings from samples with different class labels are treated as negatives. Mathematically, the positive set for anchor $\rho$ is defined as
\begin{equation}
    \mathcal{P}(\rho)
    =
    \left\{
    q \in \{1,\dots,B\} :
    q \neq \rho,\;
    y_q = y_{\rho}
    \right\},
\end{equation}
and the negative set is defined as
\begin{equation}
    \mathcal{N}(\rho)
    =
    \left\{
    a \in \{1,\dots,B\} :
    y_a \neq y_{\rho}
    \right\}.
\end{equation}

This formulation encourages ECG samples with the same abnormality status to form compact clusters in the embedding space, while separating positive and negative samples from each other.

Given an anchor embedding ${\zeta}_{\rho}$, we first compute its similarity to all other embeddings in the mini-batch. We use cosine similarity, defined as
\begin{equation}
    s_{\rho q}
    =
    \mathrm{sim}
    \left(
    {\zeta}_{\rho},
    {\zeta}_{q}
    \right)
    =
    \frac{
    {\zeta}_{\rho}^{\top}{\zeta}_{q}
    }{
    \|{\zeta}_{\rho}\|_2
    \|{\zeta}_{q}\|_2
    },
    \quad q \neq \rho .
\end{equation}

The standard supervised contrastive loss for the anchor sample $\rho$ is formulated as
\begin{equation}
    \mathcal{L}_{\mathrm{SCL}}^{(\rho)}
    =
    -\frac{1}{|\mathcal{P}(\rho)|}
    \sum_{q \in \mathcal{P}(\rho)}
    \log
    \frac{
    \exp\left(s_{\rho q}/\tau\right)
    }{
    \sum\limits_{a=1, a\neq \rho}^{B}
    \exp\left(s_{\rho a}/\tau\right)
    },
\end{equation}
where $\tau>0$ is the temperature parameter that controls the concentration of the similarity distribution. A smaller $\tau$ encourages sharper separation between positive and negative samples, whereas a larger $\tau$ produces a smoother contrastive distribution.

Although the supervised contrastive loss improves feature-level separability, all positive pairs contribute equally to the anchor loss. This may be suboptimal when many same-class pairs are already easy to align, while hard positive pairs remain insufficiently emphasized. To address this issue, we introduce a focal weighting term into the supervised contrastive objective.
For each positive pair $(\rho,q)$, we define the contrastive probability as
\begin{equation}
    \eta_{\rho q}
    =
    \frac{
    \exp\left(s_{\rho q}/\tau\right)
    }{
    \sum\limits_{a=1, a\neq \rho}^{B}
    \exp\left(s_{\rho a}/\tau\right)
    }.
\end{equation}

Then, the focal supervised contrastive loss for anchor sample $\rho$ is defined as
\begin{equation}
    \mathcal{L}_{\mathrm{FSCL}}^{(\rho)}
    =
    -\frac{1}{|\mathcal{P}(\rho)|}
    \sum_{q \in \mathcal{P}(\rho)}
    \left(1-\eta_{\rho q}\right)^{\gamma_c}
    \log
    \left(\eta_{\rho q}\right),
\end{equation}
where $\gamma_c \geq 0$ is the focusing parameter for the contrastive objective. When $\eta_{\rho q}$ is large, the positive pair is already well aligned with the anchor, and the weighting factor $(1-\eta_{\rho q})^{\gamma_c}$ reduces its contribution. In contrast, when $\eta_{\rho q}$ is small, the positive pair is difficult to align, and the focal term assigns it a larger relative contribution.

The mini-batch focal supervised contrastive loss is obtained by averaging over all valid anchors:
\begin{equation}
    \mathcal{L}_{\mathrm{FSCL}}
    =
    \frac{1}{|\mathcal{I}|}
    \sum_{\rho \in \mathcal{I}}
    \mathcal{L}_{\mathrm{FSCL}}^{(\rho)},
\end{equation}
where
\begin{equation}
    \mathcal{I}
    =
    \left\{
    \rho \in \{1,\dots,B\} :
    |\mathcal{P}(\rho)| > 0
    \right\}
\end{equation}
denotes the set of anchors that have at least one positive sample in the mini-batch.

Finally, the overall training objective combines the focal binary cross-entropy loss and the focal supervised contrastive loss:
\begin{equation}
    \mathcal{L}_{\mathrm{total}}
    =
    \mathcal{L}_{\mathrm{FBCE}}
    +
    \lambda
    \mathcal{L}_{\mathrm{FSCL}},
\end{equation}
where $\lambda$ controls the contribution of the contrastive regularization term.

By combining the classification loss and the proposed focal supervised contrastive loss, the model is optimized from two complementary perspectives. 
The classification loss focuses on improving decision boundary separability at the output level, while the contrastive loss explicitly enforces a structured embedding space by enhancing intra-class compactness and inter-class separability. 
Together, they lead to more discriminative and robust feature representations, ultimately improving overall classification performance.

\subsection{Perturbation-based Lead Importance}

In the task of myocardial substrate abnormality detection, understanding which ECG leads the model relies on for its predictions is of significant clinical importance. Myocardial substrate abnormalities often manifest as structural alterations in cardiac tissue, which can disrupt normal electrical conduction and alter ECG morphology. Since ECG signals capture the electrical activity of the heart projected along different lead directions, each lead provides a distinct spatial view of cardiac function. Therefore, analyzing lead importance helps assess whether the model captures physiologically relevant patterns associated with pathological regions.
Moreover, ECG signals are inherently multi-lead, temporally dynamic, and often exhibit strong inter-lead correlations, making model predictions difficult to interpret directly. As a result, incorporating interpretability analysis not only helps reveal the underlying decision mechanisms of the model but also enhances its reliability and trustworthiness in clinical applications. 

In the current investigation, we engage permutation importance to quantify the contribution of each input lead to the model's predictions~\cite{fisher2019all,celermajer2012cardiovascular}. 
This method is model-agnostic, not relying on the internal structure of the model. Instead, it evaluates feature importance by measuring the impact of feature perturbation on overall performance metrics, thereby characterizing the model's reliance on different leads at the task level. Consequently, permutation importance provides a more intuitive and robust assessment of feature importance. Furthermore, compared to methods based on individual samples or local gradients, this perturbation-based analysis evaluates the effect of removing lead-specific information across the test distribution, offering a more comprehensive reflection of each lead's contribution to the model's predictions.
Note that permutation importance is evaluated on a held-out test dataset to assess the contribution of each input lead without introducing bias from the training process. 

\begin{figure*}[htbp]
\centering
\includegraphics[width=1.0\linewidth]{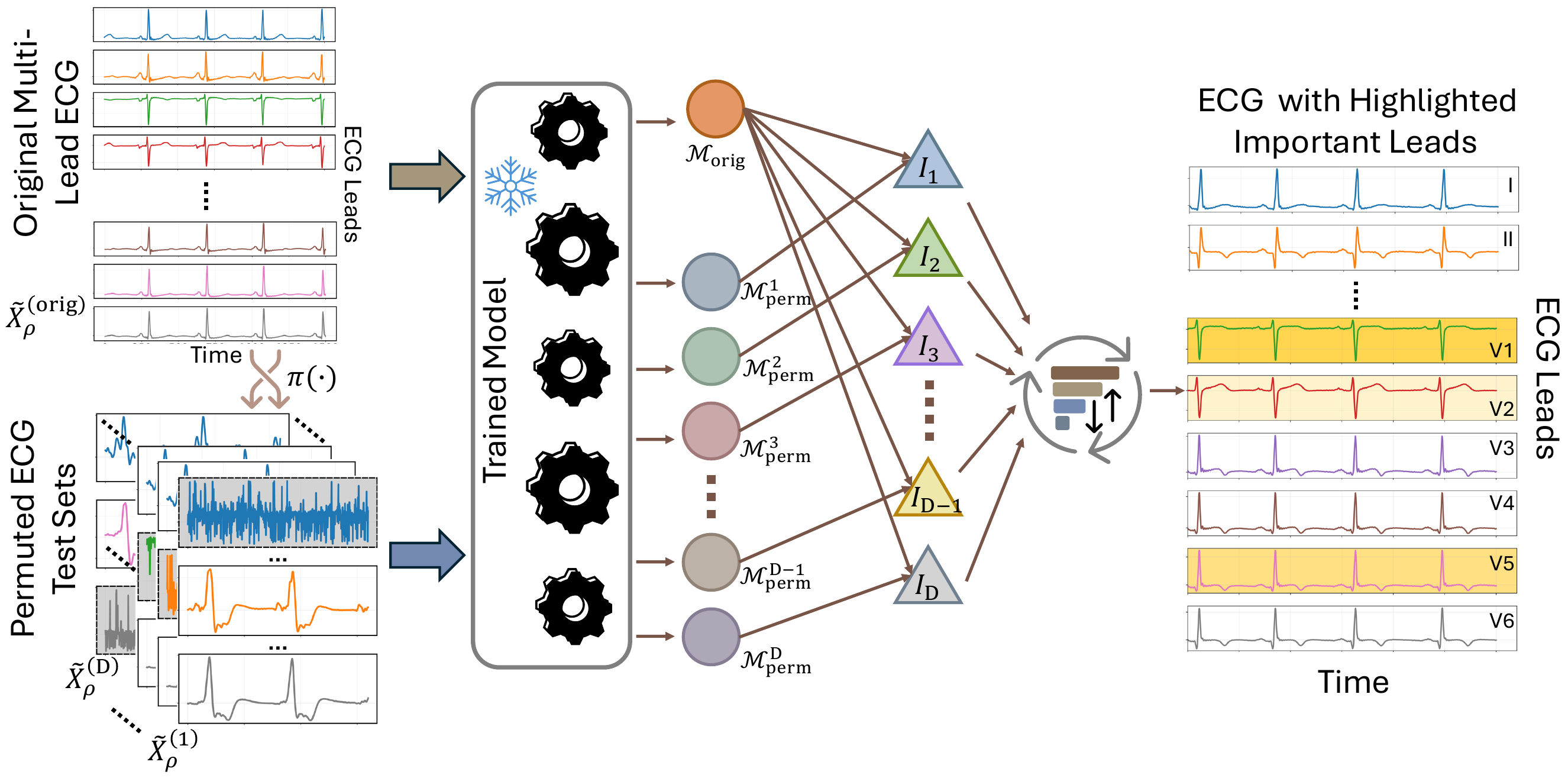}
\caption{Permutation Importance Process. }
\label{fig:pi}
\end{figure*}

The process of permutation importance is illustrated in Fig.~\ref{fig:pi}. To quantify the contribution of each ECG lead, we construct a perturbed test dataset by selectively disrupting one chosen lead while keeping all remaining leads unchanged. Specifically, for the $j$-th lead, we replace its original waveform with Gaussian noise. This operation removes the lead-specific diagnostic information. The resulting performance degradation reflects the extent to which the model relies on the disrupted lead for myocardial substrate abnormality classification.

Let the original test dataset be
\begin{equation}
    \mathcal{D}_{\mathrm{test}}
    =
    \left\{
    \left(X_{\rho}, y_{\rho}\right)
    \right\}_{\rho=1}^{M},
\end{equation}
where $X_{\rho} \in \mathbb{R}^{D \times T}$ denotes the $\rho$-th ECG sample with $D$ leads and temporal length $T$, and $y_{\rho}$ is the corresponding ground-truth label.
For the $j$-th lead, we construct a perturbed ECG sample 
$\widetilde{X}_{\rho}^{(j)}$ by replacing only the $j$-th lead with Gaussian noise:
\begin{equation}
    \widetilde{X}_{\rho,d}^{(j)}(t)
    =
    \begin{cases}
    \epsilon_{\rho,j}(t), & d=j, \\
    X_{\rho,d}(t), & d\neq j,
    \end{cases}
\end{equation}
where $\epsilon_{\rho,j}(t)$ denotes Gaussian noise sampled as $\epsilon_{\rho,j}(t) \sim \mathcal{N}(\mu_j,\sigma_j^2)$.
Here, $\mu_j$ and $\sigma_j^2$ are estimated from the $j$-th lead signals in the test set, so that the perturbed lead preserves the approximate marginal scale of the original lead while removing sample-specific temporal morphology.
The corresponding perturbed test dataset is defined as
\begin{equation}
    \mathcal{D}_{\mathrm{test}}^{(j)}
    =
    \left\{
    \left(\widetilde{X}_{\rho}^{(j)}, y_{\rho}\right)
    \right\}_{\rho=1}^{M}.
\end{equation}

The original prediction probability for test sample $\rho$ is given by
\begin{equation}
    p_{\rho}
    =
    f_{\theta}(X_{\rho}),
\end{equation}
where $f_{\theta}$ denotes the trained MSAIC-Net model. After perturbing the $j$-th lead, the corresponding prediction becomes
\begin{equation}
    \widetilde{p}_{\rho}^{(j)}
    =
    f_{\theta}\left(\widetilde{X}_{\rho}^{(j)}\right).
\end{equation}
For each sample, the influence of the $j$-th lead is measured by the change in prediction caused by the perturbation:
\begin{equation}
    I_{\rho,j}
    =
    \left|
    p_{\rho}
    -
    \widetilde{p}_{\rho}^{(j)}
    \right|.
\end{equation}

The sample-wise importance score $I_{\rho,j}$ provides an individualized explanation for the $\rho$-th ECG sample by quantifying how strongly the MSAIC-Net prediction changes when the $j$-th lead is perturbed. Therefore, $I_{\rho,j}$ can be used to identify which leads contribute most to the model decision for each individual case.

In addition to this case-specific explanation, we further quantify cohort-level lead importance by measuring the change in model performance after perturbing each ECG lead across the entire test cohort. Specifically, let $\mathcal{M}$ denotes an evaluation metric, such as AUROC or AUPRC, computed on the test set. Let 
$
\boldsymbol{y} = \{y_{\rho}\}_{\rho=1}^{M}
$
denote the ground-truth labels of all test samples, and let
$\boldsymbol{p} = \{p_{\rho}\}_{\rho=1}^{M}$
denote the predicted probabilities obtained from the original ECG inputs. After perturbing the $j$-th lead for all test samples, the corresponding predicted probabilities are denoted as
$
\widetilde{\boldsymbol{p}}^{(j)}
=
\left\{
\widetilde{p}_{\rho}^{(j)}
\right\}_{\rho=1}^{M}.
$
The original model performance is computed as
\begin{equation}
    \mathbf{M}
    =
    \mathcal{M}(\boldsymbol{p}, \boldsymbol{y}),
\end{equation}
where $\mathcal{M}(\cdot)$ denotes an evaluation metric such as AUROC and AUPRC. Similarly, the perturbed performance after modifying the $j$-th lead is defined as
\begin{equation}
    \widetilde{\mathbf{M}}^{(j)}
    =
    \mathcal{M}
    \left(
    \widetilde{\boldsymbol{p}}^{(j)},
    \boldsymbol{y}
    \right).
\end{equation}
$\widetilde{\mathbf{M}}^{(j)}$ denote the same metric computed after perturbing the $j$-th lead for all test samples. The cohort-level importance score for the $j$-th lead is then defined as
\begin{equation}
    I_j
    =
    \Delta \mathbf{M}^{(j)}
    =
    \mathbf{M}
    -
    \widetilde{\mathbf{M}}^{(j)}.
\end{equation}

The cohort-level score $I_j$ reflects the overall contribution of the $j$-th lead across the study population. A larger $I_j$ indicates that perturbing this lead more consistently induces changes in the MSAIC-Net prediction, suggesting that the lead carries more influential information for myocardial abnormality classification at the population level. Since different ECG leads reflect electrical activity from different anatomical orientations, the distribution of lead-wise importance may also provide insight into regions where myocardial substrate abnormalities are more frequently manifested in the cohort.

\section{Experiments and Results}

\subsection{Dataset and Preprocessing}

%The proposed model was trained and validated using two ECG datasets: a clinical dataset collected from the University of Virginia (UVA) Health System and the publicly available PTB-XL dataset from PhysioNet. The UVA cohort included 447 patients and 7,056 ECG recordings. Myocardial scar labels were derived from expert-annotated late gadolinium enhancement cardiac magnetic resonance (LGE-CMR) examinations, and the cohort was used as the primary dataset for ECG-based myocardial scar classification. To further evaluate the model on a related ECG abnormality detection task, we used a subset of the PTB-XL dataset for MI identification. Specifically, only ECG recordings labeled as MI or normal were included, while recordings with other diagnostic categories were excluded. This resulted in 2950 MI recordings and 9478 normal recordings from PTB-XL. Both ECG datasets were processed using a unified preprocessing pipeline before model training:

The proposed model was trained and validated using two ECG datasets. The first dataset was a myocardial scar dataset collected from the University of Virginia (UVA) Health System. 
% This dataset included 429 patient-level ECG samples, among which 117 were labeled as normal and the remaining samples were labeled as positive for myocardial scar. 
Scar labels were determined based on corresponding late gadolinium enhancement cardiac magnetic resonance imaging (LGE-CMR) examinations. Because the UVA cohort had a relatively limited number of patient-level samples, we applied a segmentation-based data augmentation strategy. Specifically, each ECG recording was divided into fixed-length temporal samples, allowing multiple training samples to be generated from one recording while preserving the original patient-level scar label. After segmentation-based augmentation, the UVA dataset contained 7,056 ECG samples, including 5,632 training samples and 1,424 testing samples, corresponding to an approximately 80\%/20\% train-test split. The training set remained imbalanced, with 3,968 scar-positive samples and 1,664 normal samples, corresponding to a positive rate of 70.5\% and a positive-to-normal ratio of approximately 2.38:1.

% The second dataset was derived from the publicly available PTB-XL dataset from PhysioNet and was used to evaluate the proposed MSAIC-Net model on a related myocardial infarction detection task~\cite{wagner2020ptb}. Specifically, we included all normal ECG recordings, consisting of 9,514 recordings, and a subset of MI-labeled recordings, consisting of 2,950 recordings. Because this PTB-XL subset already contained a sufficient number of samples, segmentation-based augmentation was not applied to this dataset.
% The second dataset was derived from the publicly available PTB-XL dataset from PhysioNet and was used to evaluate the proposed MSAIC-Net model on a related MI detection task~\cite{wagner2020ptb}. Specifically, we included all normal ECG recordings and a subset of MI-labeled recordings. After preprocessing and window generation, the PTB-XL subset contained 12,428 ECG samples, including 9,946 training samples and 2,482 testing samples, corresponding to an 80\%/20\% train-test split. The training set was class-imbalanced, with 2,366 MI-positive samples and 7,580 negative samples, corresponding to a positive rate of 23.8\% and a negative-to-positive ratio of approximately 3.20:1. Because this PTB-XL subset already contained a sufficient number of samples, segmentation-based augmentation was not applied to this dataset.
% {{\color{red} we talked about so much IMBALANCE PROBLEM as a selling point. And in your data discription, nothing about that? what is the imbalance ratio? }}

The second dataset was derived from the publicly available PTB-XL dataset from PhysioNet~\cite{wagner2020ptb} and was used to evaluate the proposed MSAIC-Net model on a related MI detection task. After label filtering and preprocessing, the PTB-XL subset contained 12,428 ECG samples, including 2,950 MI-positive samples and 9,478 negative samples. Since PTB-XL already contains a relatively large number of ECG recordings, segmentation-based augmentation was not applied to this dataset. The dataset was further split into training and testing sets using an 80\%/20\% train-test split, resulting in 9,946 training samples and 2,482 testing samples. The training set was class-imbalanced, with 2,366 MI-positive samples and 7,580 negative samples, corresponding to a positive rate of 23.8\% and a negative-to-positive ratio of approximately 3.20:1.

For both datasets, ECG signals were preprocessed using noise-filtering procedures implemented in the BioSPPy Python package to reduce baseline drift and other artifacts~\cite{bota2024biosppy}. All ECG recordings were further amplitude-normalized to reduce inter-patient variability. To avoid data leakage, all training and testing partitions were performed at the patient level, ensuring that ECG recordings or augmented segments derived from the same patient did not appear across different data splits.

% \begin{itemize}
% \item \textbf{Noise filtering:} baseline drift, calibration pulses, and segments with excessive noise or flat-line artifacts were removed using a combination of digital filtering and threshold-based rules.  
% \item \textbf{Resampling and normalization:} all ECG leads were resampled to a uniform 500~Hz sampling rate and amplitude-normalized to reduce inter-patient variability.  
% \item\textbf{Segmentation:} for the UVA cohort, given the limited number of patient samples, we augmented the training data by segmenting each ECG recording into fixed-length temporal windows. This approach increased the effective number of training samples.
% \item\textbf{Train-test split:} all data were partitioned at the \textit{patient level} into training, validation, and testing sets to avoid data leakage across segments derived from the same patient.
% \end{itemize}

% This standardized preprocessing procedure ensured consistency across subjects and minimized acquisition-related variability, enabling the deep learning models to learn physiologically meaningful waveform patterns rather than artifacts or instrumentation noise.

\subsection{Evaluation Metrics}
\label{Sec:metrics}

Model performance was evaluated using the area under the receiver operating characteristic curve (AUROC), the area under the precision--recall curve (AUPRC), and the F1-score.
Specifically, the receiver operating characteristic (ROC) curve describes the trade-off between sensitivity and specificity across different classification thresholds. AUROC is defined as the area under the ROC curve and measures the ability of the model to distinguish positive samples from negative samples over all possible thresholds. A larger AUROC indicates better overall discriminative ability, with a value of 1 representing perfect classification and a value of 0.5 corresponding to random guessing.
The precision--recall (PR) curve describes the relationship between precision and recall across different classification thresholds. AUPRC is defined as the area under the PR curve and is particularly informative when the dataset is imbalanced, since it focuses on the model's ability to correctly identify positive cases. A larger AUPRC indicates better performance in detecting positive samples while reducing false positives.

The F1-score is the harmonic mean of precision and recall, defined as
\begin{equation}
    \mathrm{F1}
    =
    2 \cdot
    \frac{\mathrm{Precision} \cdot \mathrm{Recall}}
    {\mathrm{Precision} + \mathrm{Recall}},
\end{equation}
where
\begin{equation}
    \mathrm{Precision}
    =
    \frac{\mathrm{TP}}{\mathrm{TP}+\mathrm{FP}},
    \qquad
    \mathrm{Recall}
    =
    \frac{\mathrm{TP}}{\mathrm{TP}+\mathrm{FN}}.
\end{equation}
Here, $\mathrm{TP}$, $\mathrm{FP}$, and $\mathrm{FN}$ denote true positives, false positives, and false negatives, respectively. The F1-score balances the model's ability to correctly detect positive cases and avoid false positive predictions. A larger F1-score indicates better classification performance, with a maximum value of 1.

\subsection{Model Validation and Ablation}

\subsubsection{Ablation study on UVA datasets }

\begin{table*}[htbp]
\centering
\small
\caption{Ablation study of different backbone architectures and training strategies on the UVA dataset}
\label{tab:ab_uva400}
\begin{tabular}{lccc}
\toprule
Backbone 
& AUPRC 
& AUROC 
& F1 Score\\
\midrule
\multirow{1}{*}{CNN + FBCE} 
  & 0.8935 & 0.7853 & 0.8593 \\
\multirow{1}{*}{MS-CNN + FBCE} 
  & 0.9475 & 0.8857 & 0.9095 \\
\multirow{1}{*}{MS-CNN+CA + FBCE} 
  & 0.9558 & 0.8978 & 0.9008 \\
\multirow{1}{*}{MS-CNN+CA + FBCE + SCL} 
  & 0.9634 & 0.9111 & 0.9016 \\
\multirow{1}{*}{MS-CNN+CA + FBCE + FSCL (MSAIC-Net)} 
  & 0.9757 & 0.9366 & 0.9169 \\
\bottomrule
\end{tabular}
\end{table*}

Table~\ref{tab:ab_uva400} presents the ablation results on the UVA test set. We progressively evaluate the each contribution of our proposed MSAIC-Net, i.e., multi-scale convolution (MS-CNN), adaptive channel attention (CA), and representation-level regularization through supervised contrastive learning (SCL) and its focal-weighted variant (FSCL), all built upon the baseline CNN trained with Focal BCE loss. Note that the baseline model used in the ablation study is a vanilla 1D-CNN trained with focal binary cross-entropy (FBCE) loss.

Starting from the conventional CNN backbone, incorporating the multi-scale atrous convolution yields a substantial performance improvement. AUPRC increases from 0.8935 to 0.9475, AUROC from 0.7853 to 0.8857, and F1 from 0.8593 to 0.9095. This marked gain demonstrates the importance of capturing ECG waveform morphology across multiple temporal scales. Myocardial scarring often manifests through both localized waveform distortions and broader conduction alterations. Fixed receptive fields are insufficient to capture such heterogeneous temporal patterns, whereas multi-scale atrous convolution enables the model to integrate both short-range and long-range dependencies, significantly enhancing discriminative capability.
Adding the channel attention module further improves AUPRC (0.9558) and AUROC (0.8978).
Channel attention dynamically recalibrates embedding responses, allowing the network to emphasize diagnostically informative features while suppressing noise and redundant signals introduced by inter-patient variability and acquisition differences. 
Incorporating supervised contrastive learning further refines representation quality, with AUPRC increasing to 0.9634 and AUROC to 0.9111 and F1 remains stable (0.9016). This indicates that representation-level regularization improves intra-class compactness and enlarges inter-class margins, which is particularly beneficial for distinguishing subtle morphological differences between normal and scarred ECG samples.

Replacing SCL with the proposed FSCL yields the best overall performance, achieving AUPRC of 0.9757, AUROC of 0.9366, and F1 of 0.9169. Compared with the MS-CNN+CA model, FSCL improves AUPRC by 0.0199 and AUROC by 0.0388, while also increasing F1 by 0.0161. The focal-style weighting mechanism emphasizes hard and minority-class samples during representation learning, effectively mitigating class imbalance without altering the original data distribution. The consistent improvements across all metrics confirm that combining multi-scale feature extraction, adaptive channel weighting, and imbalance-aware contrastive regularization yields the most robust myocardial scar classification performance on the UVA dataset.

% \begin{figure}[htbp]
% \centering
% \small
% \caption{Performance Comparison of Backbone Architectures on the UVA Dataset}
% \label{fig:ablation_uva}

% \begin{subfigure}[htbp]{\linewidth}
%   \centering
%   \includegraphics[width=0.92\linewidth]{Figures/ecg_ab_plots/uva/ablation_prc.png}
%   \caption{Precision--Recall Curve Comparison.}
%   \label{fig:ablation_prc_uva}
% \end{subfigure}

% \vspace{0.05cm}

% \begin{subfigure}[htbpt]{\linewidth}
%   \centering
%   \includegraphics[width=0.92\linewidth]{Figures/ecg_ab_plots/uva/ablation_roc.png} 
%   \caption{ROC Curve Comparison.}
%   \label{fig:ablation_roc_uva}
% \end{subfigure}

% \vspace{0.05cm}

% \begin{subfigure}[htbp]{\linewidth}
%   \centering
%   \includegraphics[width=0.92\linewidth]{Figures/ecg_ab_plots/uva/ablation_best_f1_bar.png} 
%   \caption{F1 Score Comparison.}
%   \label{fig:ablation_f1_uva}
% \end{subfigure}

% \end{figure}

\begin{figure}[htbp]
\centering
\small

\begin{subfigure}[t]{\linewidth}
  \centering
  \includegraphics[width=0.72\linewidth]{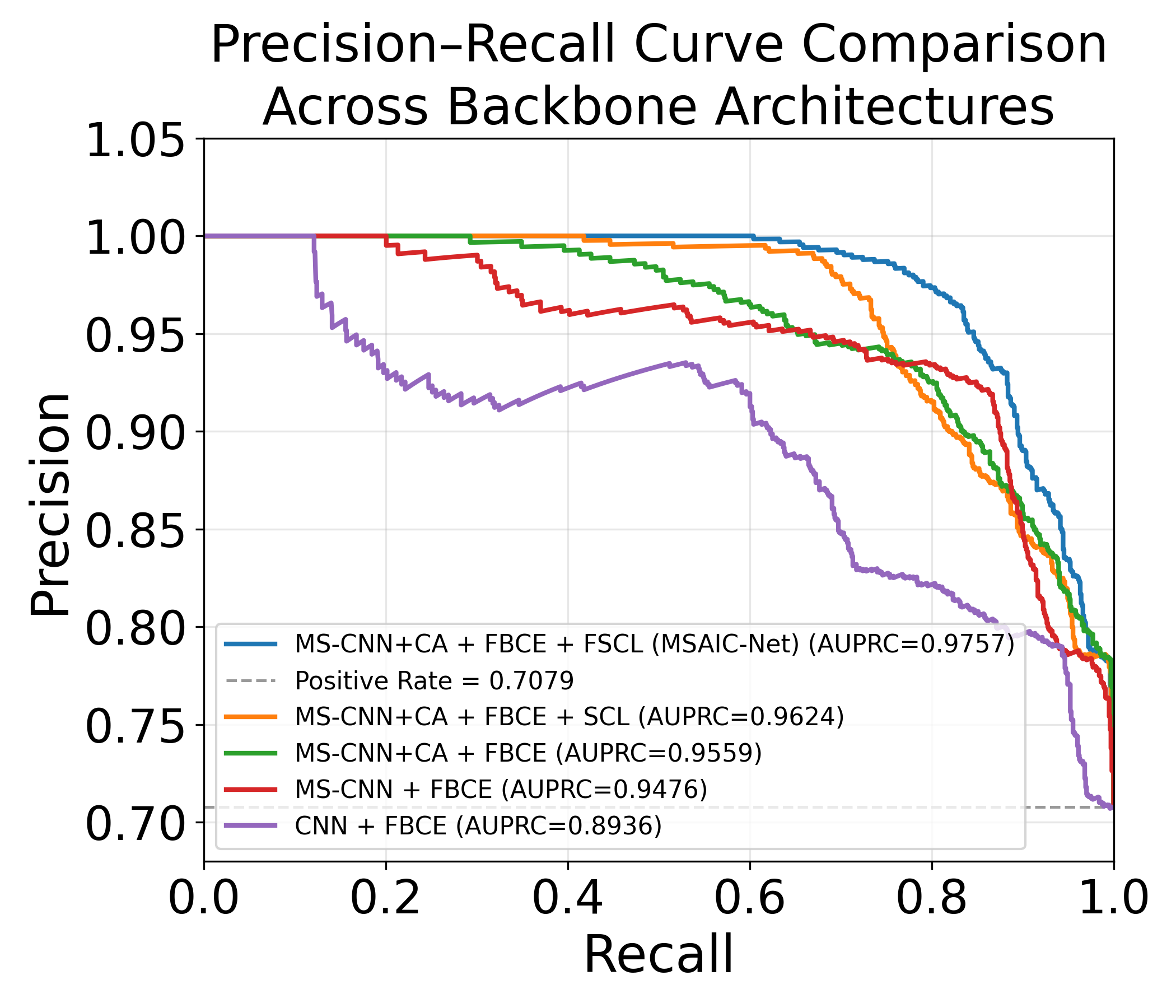}
  \caption{Precision--Recall curve comparison.}
  \label{fig:ablation_prc_uva}
\end{subfigure}

\vspace{0.05cm}

\begin{subfigure}[t]{\linewidth}
  \centering
  \includegraphics[width=0.72\linewidth]{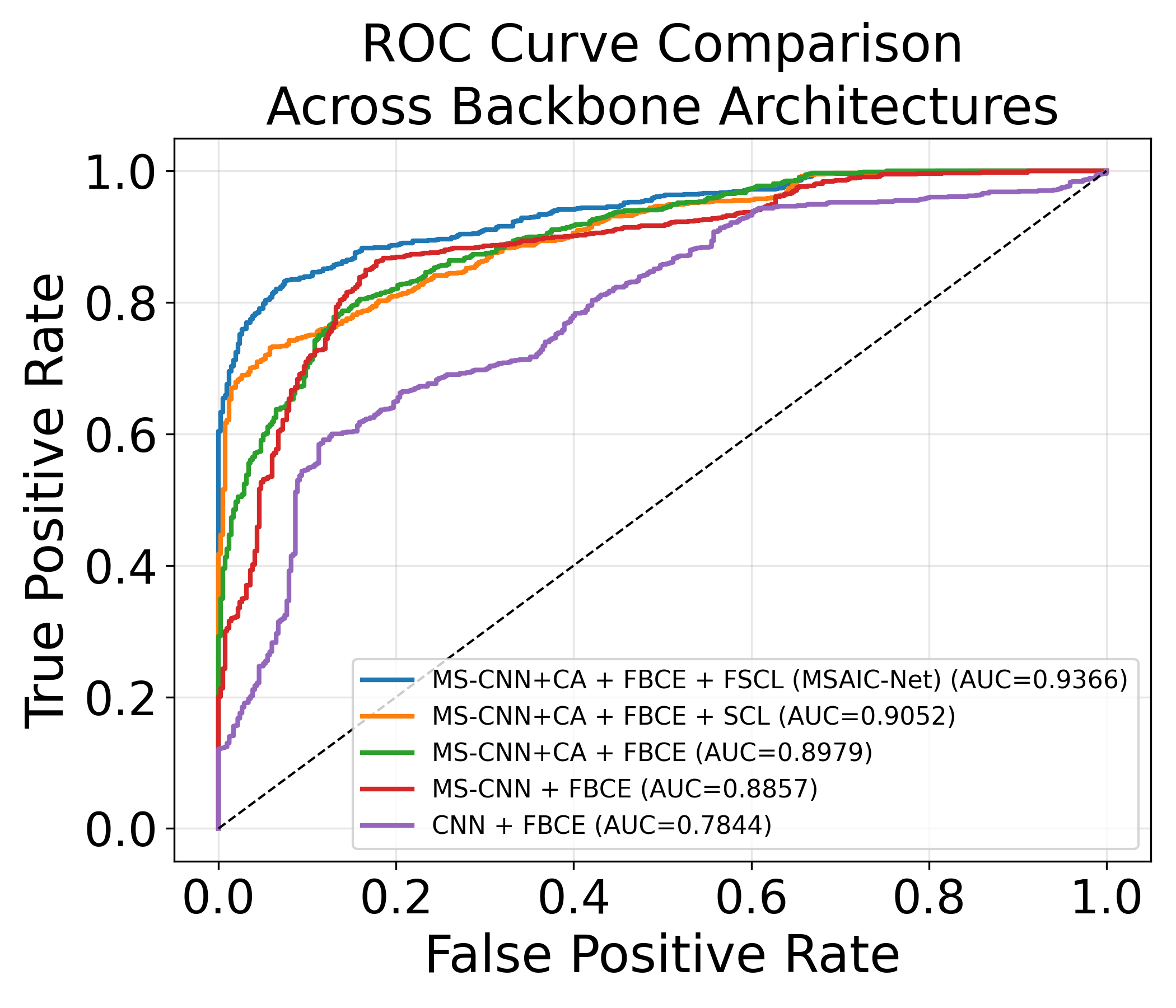} 
  \caption{ROC curve comparison.}
  \label{fig:ablation_roc_uva}
\end{subfigure}

\caption{Performance comparison of backbone architectures on the UVA dataset.}
\label{fig:ablation_uva}
\end{figure}

\begin{figure}[htbp]
\ContinuedFloat
\centering
\small

\begin{subfigure}[t]{\linewidth}
  \centering
  \includegraphics[width=0.72\linewidth]{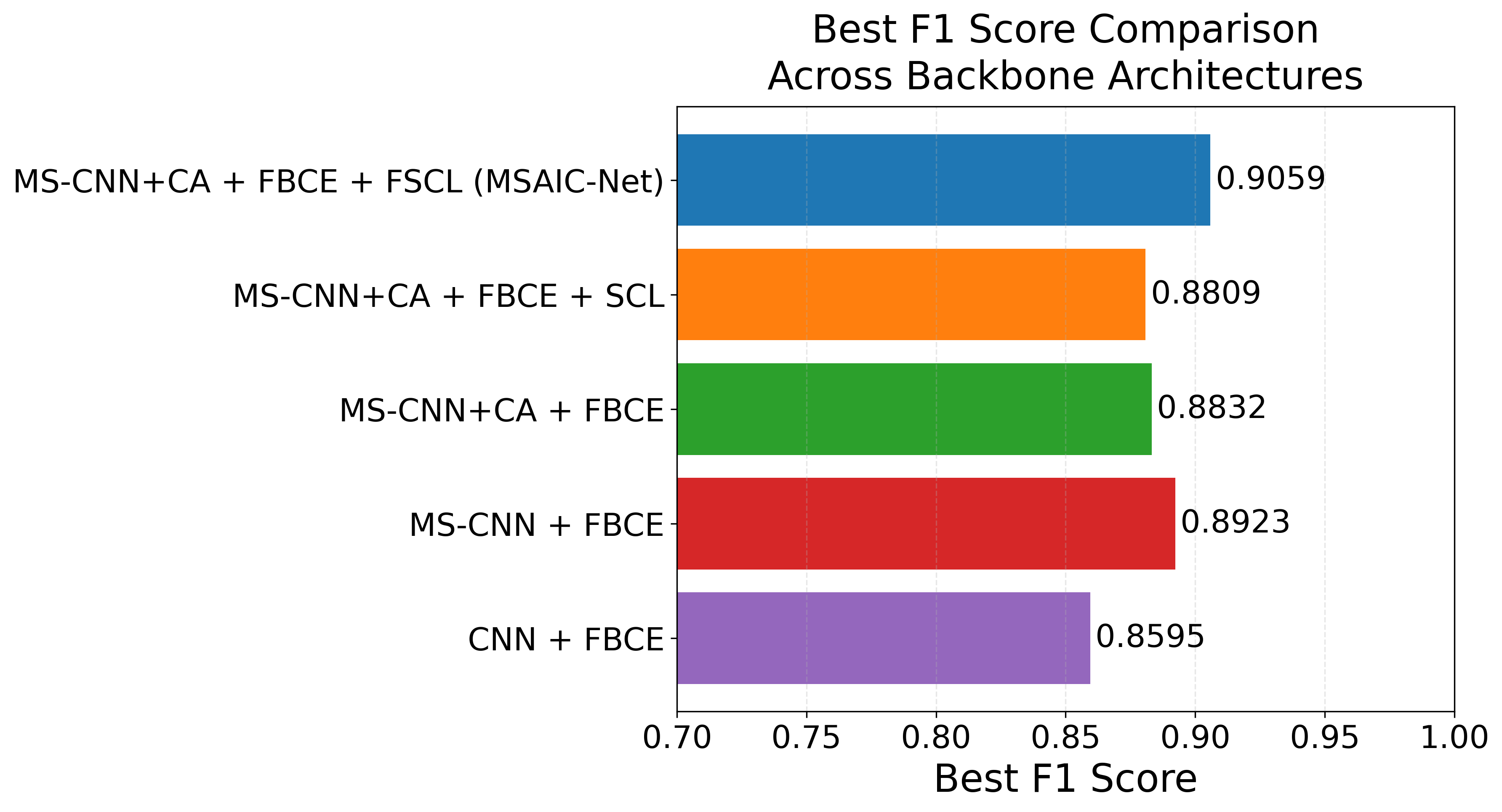} 
  \caption{F1 score comparison.}
  \label{fig:ablation_f1_uva}
\end{subfigure}

\caption{Performance comparison of backbone architectures on the UVA dataset. Continued.}
\end{figure}

Fig. ~\ref{fig:ablation_uva} presents the precision–recall curve (PRC), receiver operating characteristic curve (ROC), and F1 score comparisons across different backbone architectures on the UVA dataset.

The PR curves in Fig.~\ref{fig:ablation_prc_uva} provide further insight beyond the tabulated results. While all models exhibit decreasing precision as recall increases, the baseline CNN shows a rapid precision drop in the high-recall region, indicating limited robustness when attempting to recover more positive samples. In contrast, models with multi-scale atrous convolution and CA demonstrate a more gradual decline, suggesting improved stability across varying decision thresholds.
The proposed FSCL consistently dominates the PR curve across almost the entire recall range, indicating superior ranking capability rather than improvement at a specific operating point. The advantage is particularly pronounced in the high-recall regime, where the model maintains substantially higher precision compared to all other variants. This behavior suggests that the focal-weighted contrastive learning mechanism effectively improves the handling of hard and ambiguous samples, leading to more reliable predictions when high sensitivity is required.

The ROC curves in Fig.~\ref{fig:ablation_roc_uva} provide a visual perspective on model discrimination. While all models show improved performance compared to the baseline, the proposed method consistently achieves a curve that is closer to the top-left corner across most of the FP rate range.
In particular, the advantage is most evident in the low-FPR region, where the proposed model attains a higher TP rate under stricter FP constraints. This indicates improved sensitivity without incurring excessive false alarms, which is especially important in clinical screening scenarios.
Furthermore, the ROC curve of the proposed method remains consistently above those of other variants, suggesting a globally superior ranking of positive and negative samples rather than improvement at a specific threshold. 
The best F1 score comparison in Fig.~\ref{fig:ablation_f1_uva} shows that the proposed method achieves the highest performance at the optimal classification decision threshold.

\subsubsection{Ablation study on PTB-XL dataset}

% \begin{table*}[htpb]
% \centering
% \small
% \caption{Ablation study of different backbone architectures and training strategies on the PTB-XL dataset}
% \label{tab:ab_PTB-XL400}
% \begin{tabular}{lccc}
% \toprule
% Backbone 
% & AUPRC (test) 
% & AUROC (test) 
% & F1 Score\\
% \midrule
% \multirow{1}{*}{CNN + FBCE} 
%   & 0.9042 & 0.9531 & 0.8209 \\
% \multirow{1}{*}{MS-CNN + FBCE} 
%   & 0.9277 & 0.9653 & 0.8579 \\
% \multirow{1}{*}{MS-CNN + CA + FBCE} 
%   & 0.9299 & 0.9653 & 0.8498 \\
% \multirow{1}{*}{MS-CNN+CA + FBCE + SCL} 
%   & 0.9306 & 0.9700 & 0.8530 \\
% \multirow{1}{*}{MS-CNN+CA  + FBCE + FSCL (MSAIC-Net)} 
%   & 0.9343 & 0.9678 & 0.8576 \\
% \bottomrule
% \end{tabular}
% \end{table*}

\begin{table*}[htbp]
\centering
\small
\setlength{\tabcolsep}{4pt}
\renewcommand{\arraystretch}{0.9}
\caption{Ablation study of different backbone architectures and training strategies on the PTB-XL dataset}
\label{tab:ab_PTB-XL400}
\resizebox{\textwidth}{!}{
\begin{tabular}{p{0.48\textwidth}ccc}
\toprule
Backbone 
& AUPRC (test) 
& AUROC (test) 
& F1 Score\\
\midrule
CNN + FBCE 
  & 0.9042 & 0.9531 & 0.8209 \\
MS-CNN + FBCE 
  & 0.9277 & 0.9653 & 0.8579 \\
MS-CNN + CA + FBCE 
  & 0.9299 & 0.9653 & 0.8498 \\
MS-CNN + CA + FBCE + SCL 
  & 0.9306 & 0.9700 & 0.8530 \\
MS-CNN + CA + FBCE + FSCL (MSAIC-Net) 
  & 0.9343 & 0.9678 & 0.8576 \\
\bottomrule
\end{tabular}
}
\end{table*}

Table~\ref{tab:ab_PTB-XL400} summarizes the ablation results on the PTB-XL test set. We progressively evaluated the contributions of multi-scale atrous convolution, CA, and representation-level regularization through supervised contrastive learning for MI detection.

Compared with the baseline CNN trained with FBCE, incorporating the multi-scale atrous convolution leads to consistent performance improvements. AUPRC increases from 0.9042 to 0.9277, AUROC from 0.9531 to 0.9653, and F1 from 0.8209 to 0.8579. These gains confirm that multi-scale temporal feature extraction enhances the model’s ability to capture heterogeneous waveform morphologies. 
Adding the channel attention further improves AUPRC to 0.9299 while maintaining AUROC at 0.9653. The F1-score shows a modest decrease from 0.8579 to 0.8498, which may be attributed to changes in probability calibration. Channel attention dynamically reweights embedding channels, enabling the model to suppress noisy or redundant responses and emphasize diagnostically informative features across leads. 
Incorporating supervised contrastive learning further refines representation structure. With SupCon, AUROC increases to 0.9700 and AUPRC to 0.9306, while F1 improves to 0.8530. This indicates enhanced inter-class separability in the embedding space, where samples from the same class become more compact and distinct from other classes. 

Replacing SCL with the proposed FSCL yields the most balanced overall performance, achieving AUPRC of 0.9343, AUROC of 0.9678, and F1 of 0.8576. Compared with the MS-CNN+CA backbone, FSCL improves AUPRC by 0.0044 and F1 by 0.0078. The focal-style weighting mechanism prioritizes hard and minority-class samples during contrastive representation learning, allowing the model to enhance discriminative power without altering the intrinsic data distribution. 

% \begin{figure}[htbp]
% \centering
% \small
% \caption{Performance Comparison of Backbone Architectures on the PTB-XL Dataset}
% \label{fig:ablation_PTB-XL}

% \begin{subfigure}[htbp]{\linewidth}
%   \centering
%   \includegraphics[width=0.6\linewidth]{Figures/ecg_ab_plots/PTB-XL/ablation_prc.png}
%   \caption{Precision--Recall Curve Comparison.}
%   \label{fig:ablation_prc_PTB-XL}
% \end{subfigure}

% \vspace{0.01cm}

% \begin{subfigure}[htbp]{\linewidth}
%   \centering
%   \includegraphics[width=0.6\linewidth]{Figures/ecg_ab_plots/PTB-XL/ablation_roc.png} 
%   \caption{ROC Curve Comparison.}
%   \label{fig:ablation_roc_PTB-XL}
% \end{subfigure}

% \vspace{0.01cm}

% \begin{subfigure}[t]{\linewidth}
%   \centering
%   \includegraphics[width=0.6\linewidth]{Figures/ecg_ab_plots/PTB-XL/ablation_best_f1_bar.png} 
%   \caption{F1 Score Comparison.}
%   \label{fig:ablation_f1_PTB-XL}
% \end{subfigure}

% \end{figure}
\begin{figure}[htbp]
\centering

\begin{subfigure}[t]{\linewidth}
  \centering
  \includegraphics[width=0.72\linewidth]{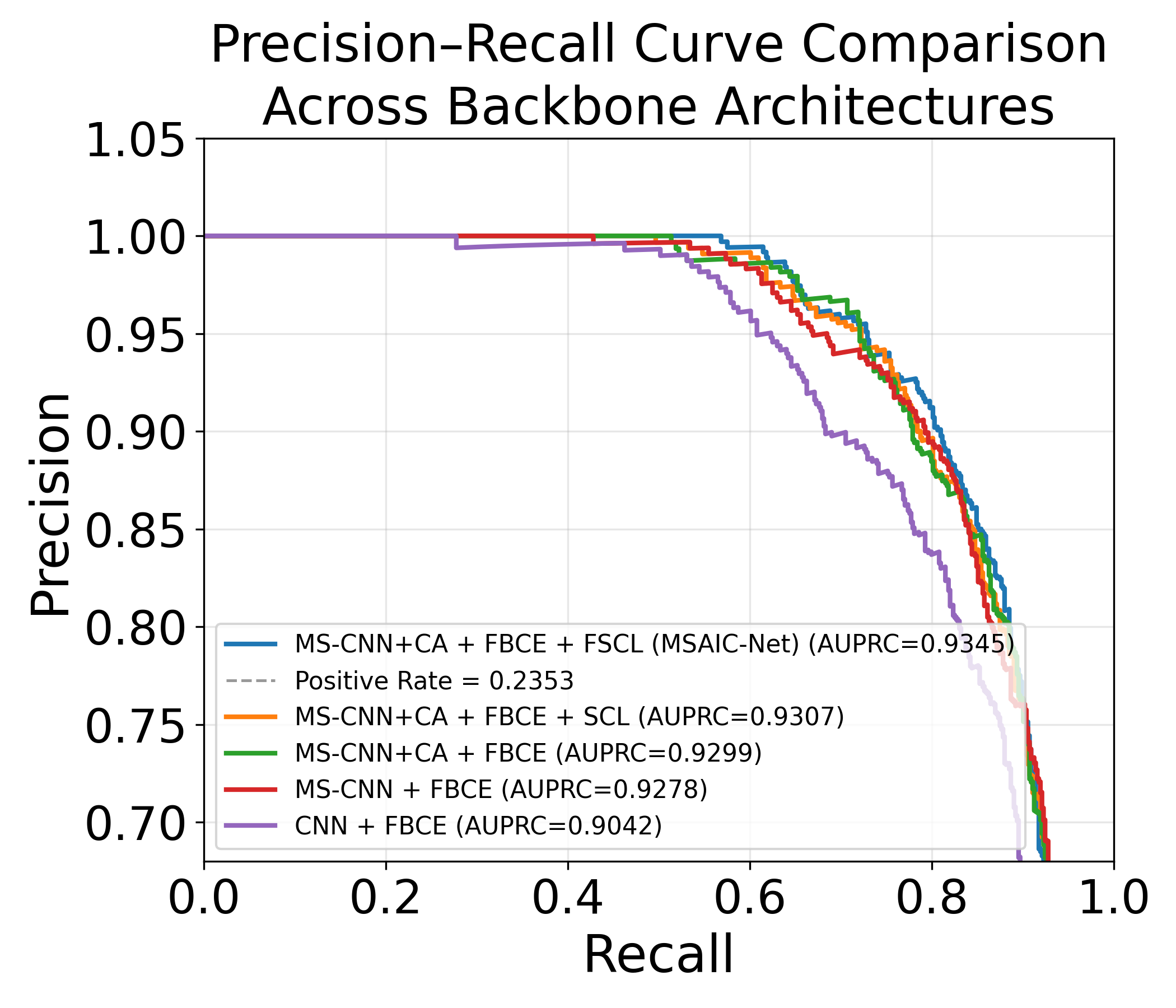}
  \caption{Precision--Recall curve comparison.}
  \label{fig:ablation_prc_PTB-XL}
\end{subfigure}

\vspace{0.1cm}

\begin{subfigure}[t]{\linewidth}
  \centering
  \includegraphics[width=0.72\linewidth]{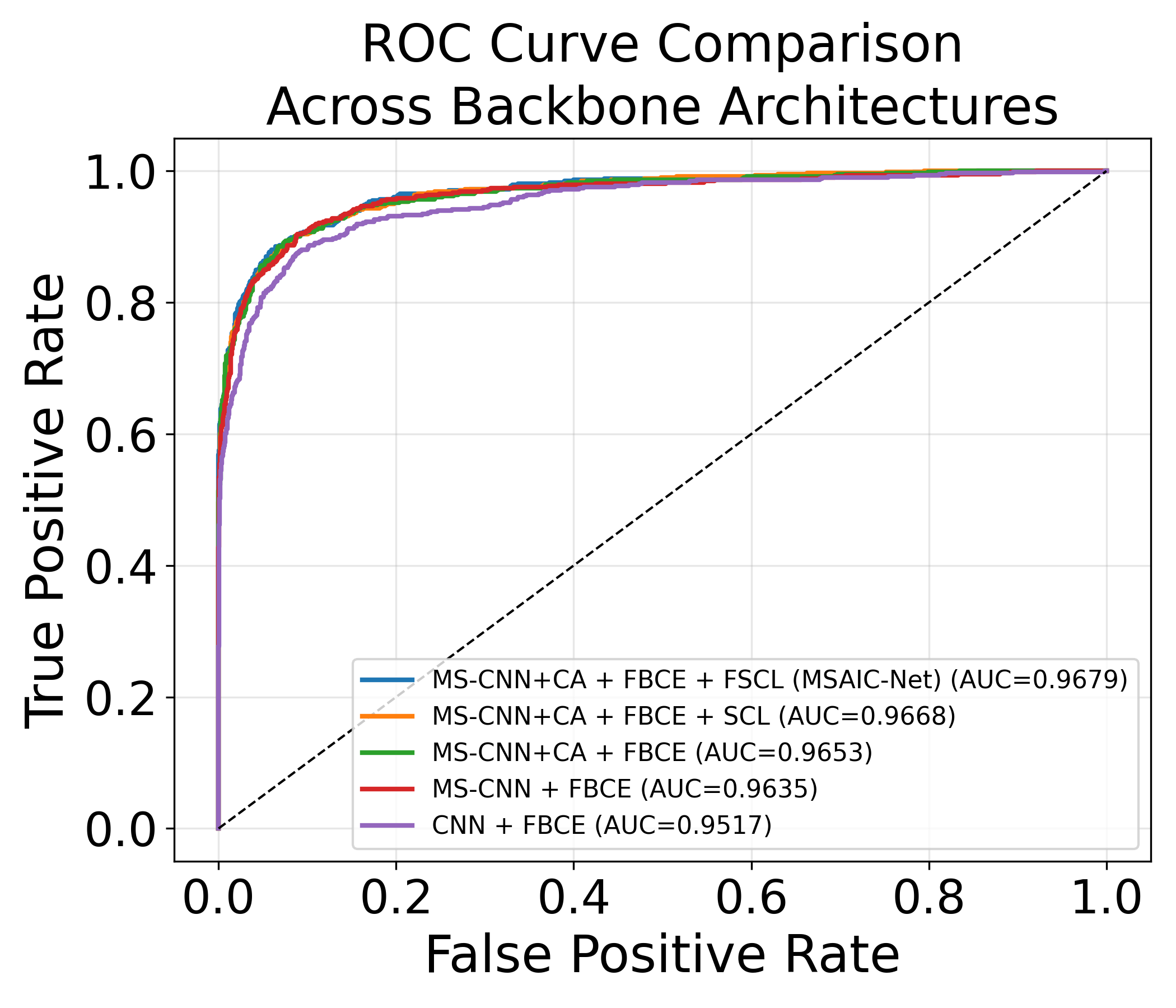} 
  \caption{ROC curve comparison.}
  \label{fig:ablation_roc_PTB-XL}
\end{subfigure}

\caption{Performance comparison of backbone architectures on the PTB-XL dataset.}
\label{fig:ablation_PTB-XL}
\end{figure}

\begin{figure}[htbp]
\ContinuedFloat
\centering

\begin{subfigure}[t]{\linewidth}
  \centering
  \includegraphics[width=0.72\linewidth]{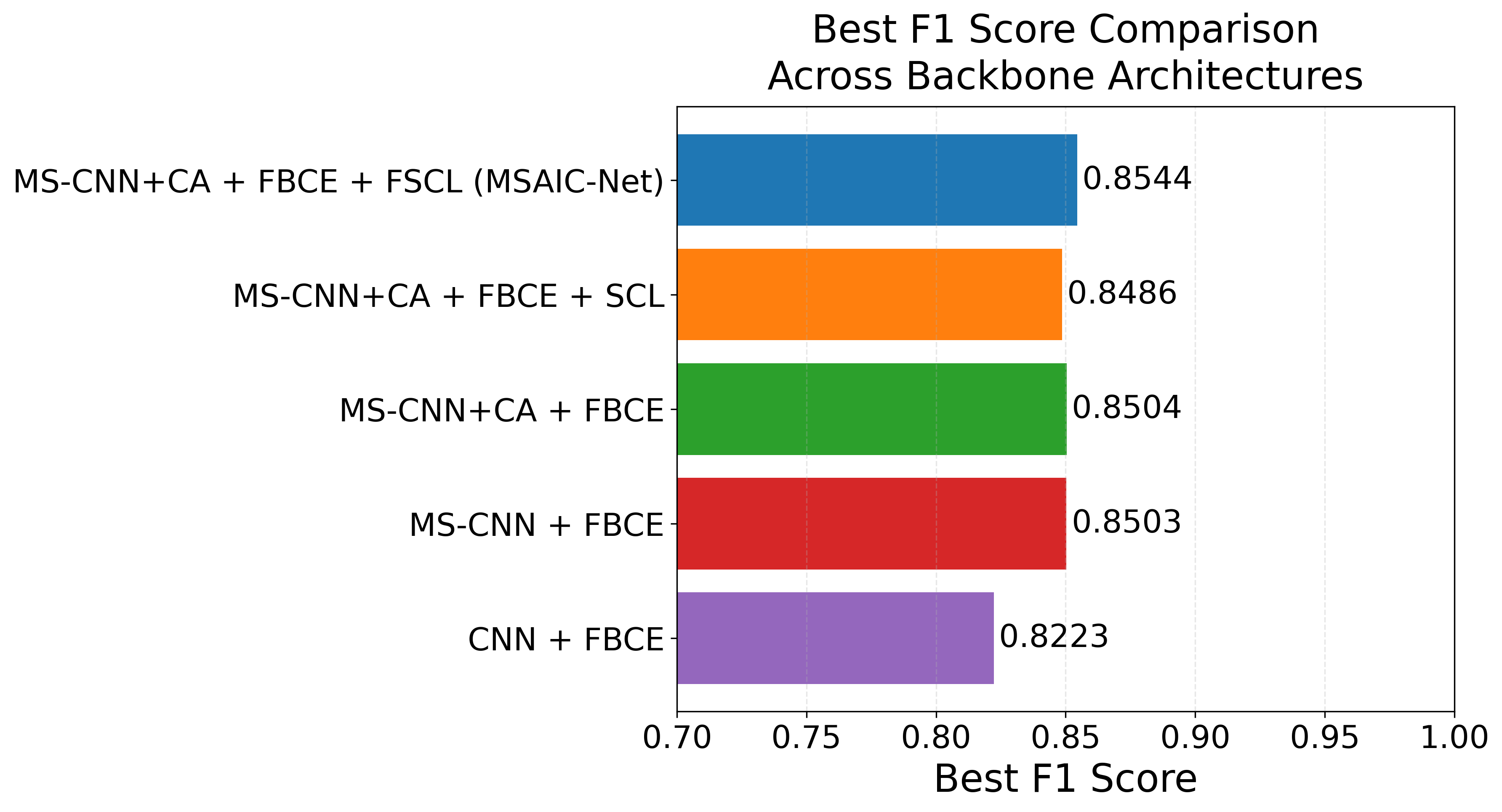} 
  \caption{F1 score comparison.}
  \label{fig:ablation_f1_PTB-XL}
\end{subfigure}

\caption{Performance comparison of backbone architectures on the PTB-XL dataset. Continued.}
\end{figure}

Fig.~\ref{fig:ablation_PTB-XL} presents the ROC curves, PR curves, and F1-score comparisons for the PTB-XL ablation study.
The PR curves on the PTB-XL dataset, shown in Fig.~\ref{fig:ablation_prc_PTB-XL}, exhibit a compact distribution, with all models achieving relatively high precision across most of the recall range. The baseline CNN still performs worse than the other variants, but the performance gap is less pronounced, suggesting that the task is inherently more stable under this dataset.
Incorporating multi-scale atrous convolution and CA leads to consistent but modest improvements, as reflected by the slightly upward-shifted curves. 
The proposed FSCL consistently achieves the highest PR curve across most of the recall range. The advantage is particularly visible in the medium-to-high recall region, where it maintains higher precision compared to other models. 

The ROC curves on the PTB-XL dataset, shown in Fig.~\ref{fig:ablation_roc_PTB-XL}, indicate that all models achieve strong discriminative performance, with curves tightly clustered near the upper-left corner. 
Despite the overall high performance, a consistent trend of improvement can still be observed. The baseline CNN yields the lowest AUC (0.9517), while the introduction of multi-scale atrous convolution and CA leads to gradual gains, reflected by the slightly upward shift of the ROC curves.
Further incorporating representation-level regularization through SCL and FSCL provides additional improvements. In particular, the proposed FSCL achieves the highest AUC (0.9679), consistently outperforming all other variants across most operating regions.
The best F1 score comparison on the PTB-XL dataset, shown in Fig.~\ref{fig:ablation_f1_PTB-XL}, reflects that MSAIC-Net provides a clear improvement over the ablation models. 

Overall, the PTB-XL ablation results are consistent with the findings from the UVA cohort. MSAIC-Net provides the primary structural improvement and imbalance-aware contrastive regularization further enhances representation separation. It is worth noting that the performance gain of the proposed method is more pronounced in the low-data setting, as observed in the UVA cohort. This suggests that the proposed architecture and regularization strategy may be particularly beneficial when the available training data are limited.

\subsection{Results of Permutation Importance}

\subsubsection{Permutation Importance Analysis on UVA dataset.}

\begin{figure}[htpb]
\centering
\small
\caption{Lead-wise permutation importance across ECG leads on the UVA dataset. The importance of each lead is quantified by the performance drop after permutation, measured in terms of (a) $\Delta \text{AUPRC}$ and (b) $\Delta \text{AUROC}$.}
\label{fig:pi_uva}

\begin{subfigure}[htpb]{\linewidth}
  \centering
  \includegraphics[width=0.92\linewidth]{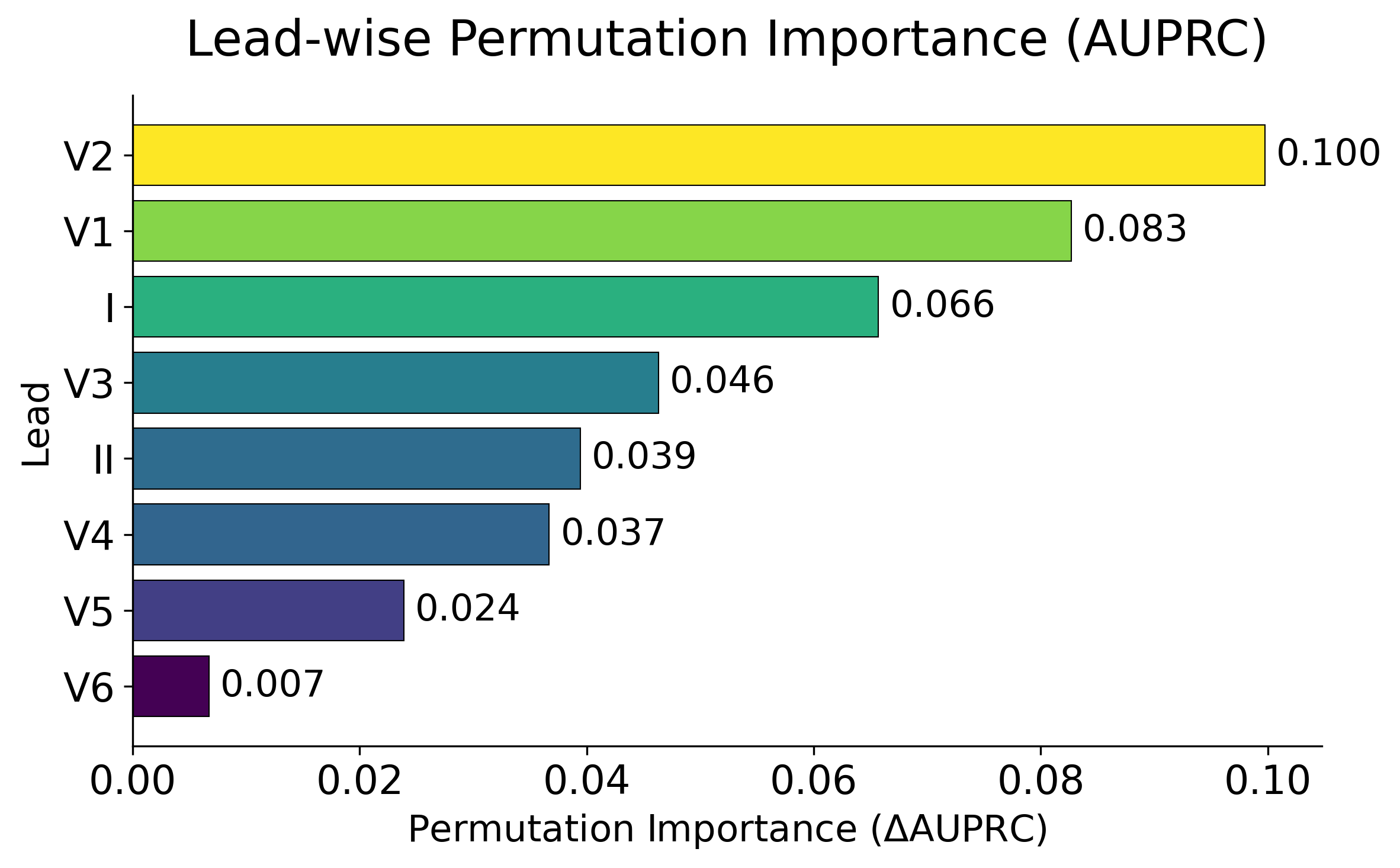}
  \caption{Lead-wise permutation importance ($\Delta$AUPRC) on UVA Dataset.}
  \label{fig:pi_prc_uva}
\end{subfigure}

\vspace{0.05cm}

\begin{subfigure}[htpb]{\linewidth}
  \centering
  \includegraphics[width=0.92\linewidth]{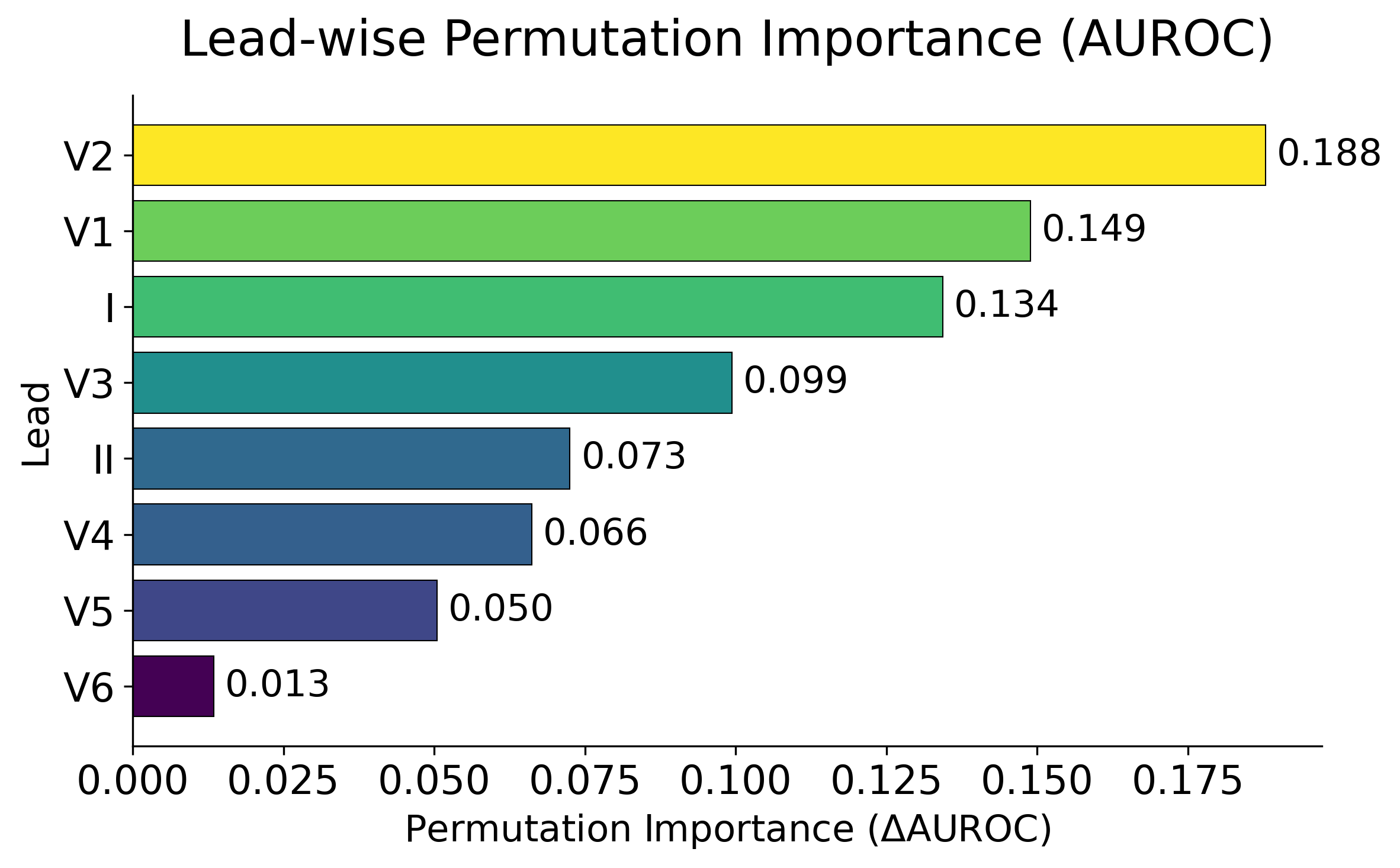} 
  \caption{Lead-wise permutation importance across seeds (AUROC) on the UVA dataset.}
  \label{fig:pi_roc_uva}
\end{subfigure}

\end{figure}

We evaluate the contribution of individual ECG leads using permutation importance. For each lead, its signal is randomly shuffled across the test set while all other leads remain unchanged, and the resulting performance drop (i.e., $\Delta$AUPRC and $\Delta$AUROC) is used as the lead importance score. The results obtained from the final trained model are shown in Fig.~\ref{fig:pi_uva}.

Note that only eight independent leads (I, II, and V1--V6) are included in the UVA dataset, while the remaining four limb leads can be derived from Leads I and II and are therefore omitted from the analysis. 

The results indicate that the model relies unevenly on different ECG leads.
Specifically, Lead V2 shows the largest performance drop under both metrics, suggesting that it provides important discriminative information for myocardial scar detection in the UVA cohort. Leads V1 and I also show relatively high importance scores. 
In contrast, Leads V3, II, and V4 exhibit moderate contributions, while Leads V5 and V6 show smaller performance drops, suggesting a relatively weaker influence on the model prediction.
Overall, the permutation importance analysis suggests that the final trained model mainly leverages a subset of informative ECG leads, while still benefiting from the multi-lead ECG representation.

\subsubsection{Permutation Importance Analysis on PTB-XL dataset}

\begin{figure}[htpb]
\centering
\small
\caption{Lead-wise permutation importance across ECG leads on the PTB-XL dataset. The importance of each lead is quantified by the performance drop after permutation, measured in terms of (a) $\Delta$AUPRC and (b) $\Delta$AUROC.}
\label{fig:pi_PTB-XL}

\begin{subfigure}[htpb]{\linewidth}
  \centering
  \includegraphics[width=0.92\linewidth]{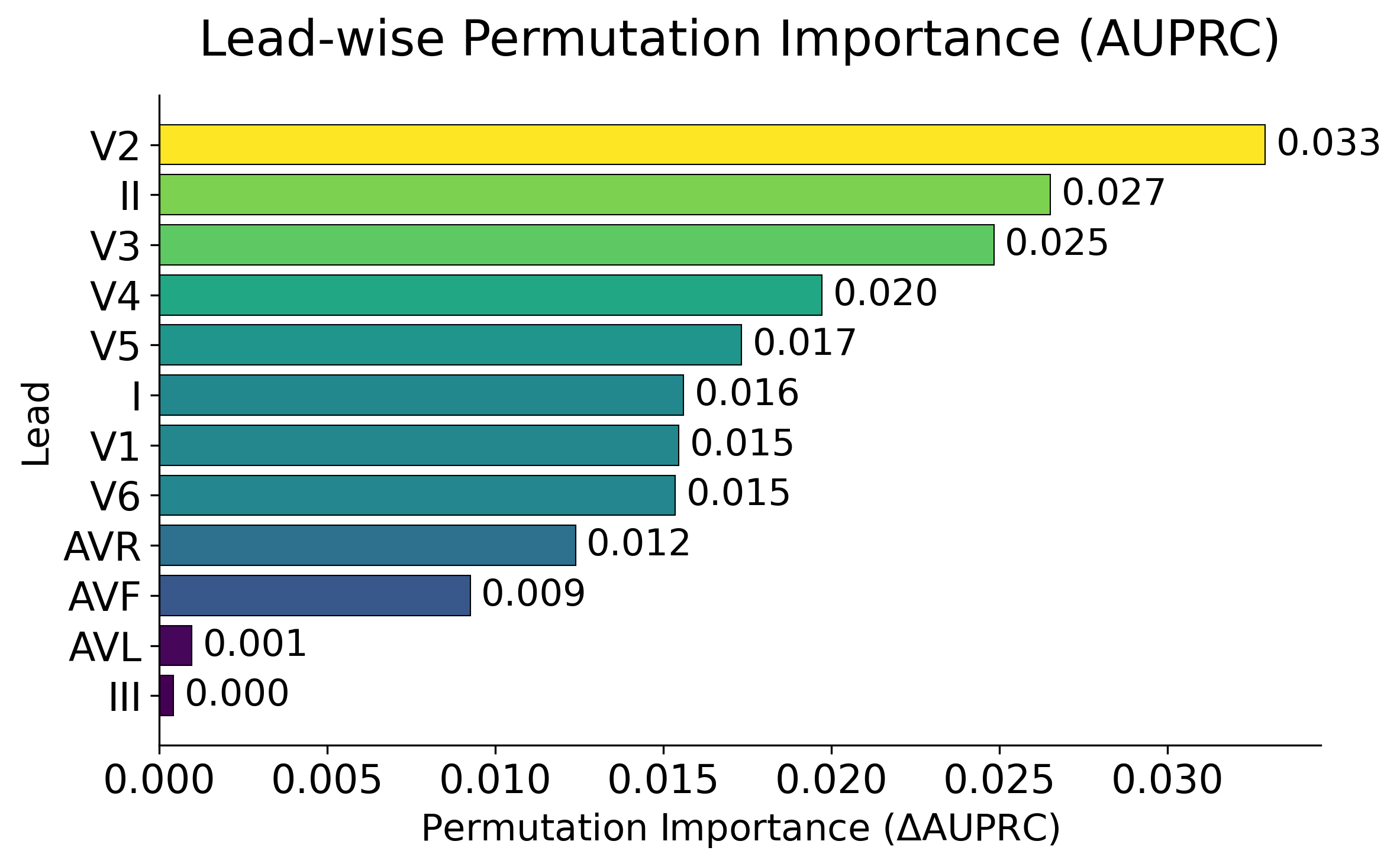}
  \caption{Lead-wise permutation importance across seeds (AUPRC) on the PTB-XL dataset.}
  \label{fig:pi_prc_PTB-XL}
\end{subfigure}

\vspace{0.1cm}

\begin{subfigure}[htpb]{\linewidth}
  \centering
  \includegraphics[width=0.92\linewidth]{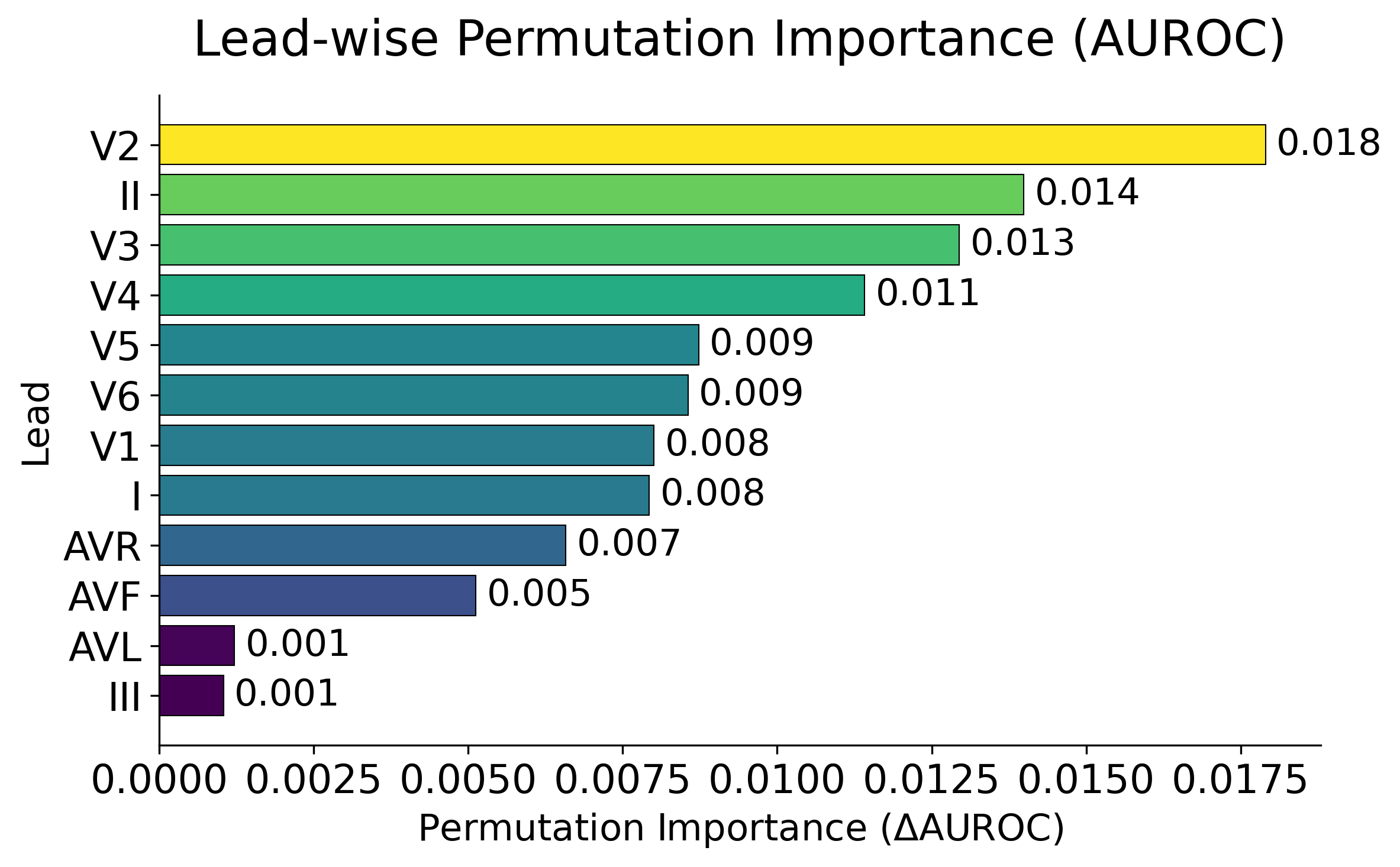} 
  \caption{Lead-wise permutation importance across seeds (AUROC) on the PTB-XL dataset.}
  \label{fig:pi_roc_PTB-XL}
\end{subfigure}

\end{figure}

We further evaluate the contribution of individual ECG leads on the PTB-XL dataset for MI detection using permutation importance. The lead-wise importance scores, measured by the performance drop ($\Delta$AUPRC and $\Delta$AUROC) after shuffling each lead, are shown in Fig.~\ref{fig:pi_PTB-XL}.

The results again indicate that the model relies unevenly on different ECG leads. In particular, Lead V2 shows the largest performance drop under both metrics, suggesting that it provides important discriminative information for MI detection. Leads V3, II, and V4 also exhibit relatively high importance scores. In contrast, Leads V5, V6, V1, I, AVR, and AVF show moderate importance, while Leads III and AVL exhibit smaller performance drops, suggesting a relatively weaker influence on the model prediction.

Overall, the permutation importance analysis suggests that the final trained model primarily leverages a subset of informative ECG leads while still benefiting from the complementary information provided by the full 12-lead ECG representation.

\subsubsection{clinical wise analysis}

In both myocardial scar classification and MI identification, each ECG lead carries distinct physiological information because it reflects cardiac electrical activity from different anatomical projections~\cite{khunti2014accurate,wagner2013marriott}. Limb leads primarily capture the frontal-plane cardiac electrical axis, whereas precordial leads V1--V6 characterize electrical activity in the horizontal plane. In general, V1--V2 are associated with septal activity, V3--V4 with anterior activity, and V5--V6 with lateral left ventricular activity~\cite{khan2004clinical,wagner2013marriott}. Therefore, abnormalities involving the septal--anterior region, particularly those related to left anterior descending (LAD) coronary artery territory, may be reflected in precordial leads such as V2. This interpretation is also supported by prior ECG-CMR studies showing that QS or Q-wave patterns in V1--V2 are associated with septal myocardial scar and that Q-wave patterns extending from V1--V2 to V3/V4 often correspond to mid-to-apical anterior, anteroseptal, or anteroapical infarction rather than isolated basal septal infarction~\cite{bogaty2002anteroseptal,allencherril2018appropriateness,ghadban2018qs}.

The relatively high cohort-level importance of Lead V2 observed in both the UVA scar classification task and the PTB-XL MI identification task is therefore clinically plausible. For MI detection, this pattern is consistent with the known involvement of V2 in septal--anterior infarction-related ECG changes. For myocardial scar, the interpretation is more indirect, since scar represents a structural substrate and ECG leads measure projected electrical activity rather than direct anatomical location~\cite{doltra2013emerging,cluitmans2015noninvasive,sattar2023electrocardiogram}. Thus, the elevated importance of Lead V2 should be interpreted as a cohort-level association with septal--anterior ECG features relevant to scar and MI, rather than direct localization of myocardial scar or infarction.

\subsection{Comparative Study}

To further evaluate the effectiveness of the proposed MSAIC-Net, we conducted a comparison study against several commonly used deep learning models for ECG classification. The goal of this experiment was to assess whether the proposed architecture provides improved predictive performance compared with standard baseline models widely used for one-dimensional ECG time-series analysis. All comparison models were trained and evaluated using the same data partitioning strategy, preprocessing pipeline, and evaluation metrics to ensure a fair comparison.

\subsubsection{Comparison on the UVA Dataset}

\begin{table}[htpb]
\centering
\caption{Comparison study on the UVA dataset}
\label{tab:cp_uva}

\adjustbox{max width=\textwidth}{
\begin{tabular}{lcccc}
\toprule
\textbf{Model} & \textbf{AUROC} & \textbf{AUPRC} & \textbf{F1}  \\
\midrule

Transformer(Ikram et al.)  & 0.7903 & 0.8905 & 0.8430 \\

Resnet(Khan et al.)       & 0.8778 & 0.9378 & 0.8839 \\

LSTM(Zhang et al.)         & 0.6954 & 0.8262 & 0.8384 \\

CNN-LSTM(Alamatsaz et al.) & 0.6094 & 0.8023 & 0.8289 \\

MSAIC-Net                  & 0.9366 & 0.9757 & 0.9169 \\

\bottomrule
\end{tabular}
}

\end{table}

Table~\ref{tab:cp_uva} presents the comparative study on the UVA dataset, where the proposed framework is evaluated against several representative ECG classification methods reported in the literature, including the Transformer-based approach proposed by Ikram et al.~\cite{ikram2025transformer}, the ResNet-based model of Khan et al.~\cite{khan2023ecg}, the LSTM-based method of Zhang et al.~\cite{zhang2019new}, and the CNN-LSTM framework developed by Alamatsaz et al.~\cite{alamatsaz2024lightweight}.

The Transformer model achieves AUROC of 0.7903 and AUPRC of 0.8905. Although Transformer-based architectures have demonstrated strong potential in large-scale ECG pretraining scenarios, their performance in this setting is constrained by dataset scale and task-specific fine-tuning requirements. Because the UVA cohort has a limited sample size, Transformer-based models may be prone to unstable representation learning and may not fully capture clinically meaningful ECG patterns~\cite{ikram2025transformer}.
ResNet based model provides improved performance compared to Transformer and recurrent architectures, achieving AUROC of 0.8778 and AUPRC of 0.9378. The convolutional inductive bias enables effective hierarchical feature extraction from ECG waveforms. However, the fixed receptive field limits its ability to capture heterogeneous temporal dependencies associated with diffuse scar-related conduction changes.
LSTM demonstrates comparatively lower performance. The LSTM model achieves AUROC of 0.6954, while the CNN-LSTM hybrid achieves AUROC of 0.6094. CNN-LSTM achieves lower performance than LSTM, indicating that increasing model complexity may bring adverse effect under a limited-data setting. Although RNNsare designed for sequential modeling, they often encounter optimization challenges when processing long ECG sequences, including gradient instability and inefficient training. 

In contrast, the proposed framework achieves the best overall performance, with AUROC of 0.9366, AUPRC of 0.9757, and F1 of 0.9169. Compared with ResNet based model, our method improves AUROC by 0.0588 and AUPRC by 0.0379, demonstrating substantial gains in both global separability and precision–recall behavior. The large improvement in AUPRC is particularly important under class imbalance, confirming the effectiveness of the focal-style contrastive regularization in enhancing minority-class representation learning.
These results indicate that multi-scale temporal modeling, adaptive channel recalibration, and imbalance-aware representation learning collectively provide superior robustness for myocardial scar classification. The consistent improvement across AUROC, AUPRC, and F1 further demonstrates that the proposed framework achieves both strong ranking performance and stable decision boundaries on the UVA dataset.

\subsubsection{Comparison on the PTB-XL Dataset}

\begin{table}[htpb]
\centering
\caption{Comparison study on the PTB-XL dataset}
\label{tab:cp_PTB-XL}

\adjustbox{max width=\textwidth}{
\begin{tabular}{lcccc}
\toprule
\textbf{Model} & \textbf{AUROC} & \textbf{AUPRC} & \textbf{F1}  \\
\midrule

Transformer(Ikram et al.)    &  0.9556 & 0.9062 & 0.7956 \\

Resnet(Khan et al.)         &  0.9635 & 0.9271 & 0.8395 \\

LSTM(Zhang et al.)           & 0.9406  & 0.8741 & 0.7618  \\

CNN-LSTM(Alamatsaz et al.)   &  0.9677 & 0.9326 & 0.8523 \\

MSAIC-Net                    &  0.9678 & 0.9343 & 0.8576 \\
\bottomrule
\end{tabular}
}

\end{table}

Table~\ref{tab:cp_PTB-XL} presents the comparative results on the PTB-XL dataset. The proposed framework is compared with several representative ECG classification methods reported in the literature, including the Transformer-based model of Ikram et al.~\cite{ikram2025transformer}, the ResNet-based model of Khan et al.~\cite{khan2023ecg}, the LSTM-based method of Zhang et al.~\cite{zhang2019new}, and the CNN-LSTM framework proposed by Alamatsaz et al.~\cite{alamatsaz2024lightweight}.
The performance obtained on PTB-XL is generally higher than that observed on the UVA dataset, which may be related to differences in dataset size, class distribution, and task complexity.
The Transformer model achieves AUROC of 0.9556 and AUPRC of 0.9062. 
ResNet based model achieves AUROC of 0.9635 and AUPRC of 0.9271, demonstrating strong performance due to its hierarchical convolutional feature extraction. The CNN-LSTM hybrid further improves AUROC to 0.9677 and AUPRC to 0.9326, indicating that combining convolutional encoding with sequential modeling enhances temporal aggregation.
The proposed framework achieves AUROC of 0.9678, AUPRC of 0.9343, and F1 of 0.8576, delivering the best overall AUPRC and F1 among all compared methods. Although the AUROC improvement over CNN-LSTM is marginal (0.9678 vs. 0.9677), the increase in AUPRC and F1 demonstrates improved precision–recall trade-off and more stable classification boundaries. Compared with ResNet, our method improves AUPRC by 0.0072 and F1 by 0.0181, confirming the benefit of multi-scale temporal modeling, adaptive channel recalibration, and imbalance-aware contrastive regularization.

Notably, all models achieve better overall performance on PTB-XL than on the UVA cohort, and the performance gaps among different comparison models are relatively small. This may be attributed to the larger sample size of PTB-XL, which provides more diverse training examples and reduces the difficulty of learning stable ECG representations compared with the limited-data UVA cohort.

Overall, the comparative results on PTB-XL corroborate the findings from the UVA dataset, demonstrating that the proposed method generalizes well across datasets with different population characteristics and acquisition conditions.

\section{Conclusions}

In this paper, we propose a multi-scale attention-enhanced convolutional network (MSAIC-Net) for myocardial substrate abnormality detection from multi-lead ECG signals. The proposed model integrates multi-scale attention-based feature reweighting and representation-level regularization to learn clinically relevant and fine-grained ECG representations associated with myocardial scar and MI-related abnormalities, contributing to improved classification performance. Furthermore, a novel focal-weighted supervised contrastive learning strategy is designed to enhance intra-class compactness and inter-class separability in the feature space, thereby alleviating the impact of class imbalance.
Experimental results demonstrate that MSAIC-Net maintains strong performance and stability under class-imbalanced conditions. It outperforms state-of-the-art models, with particularly significant improvements in the low-data regime. In addition, by incorporating interpretability analysis, the model is able to provide clinically meaningful lead-level contribution information. %, which improves its transparency and reliability.
Overall, the proposed method provides an effective and interpretable framework for ECG-based myocardial scar classification and MI identification. The method shows strong potential for large-scale screening, assisted diagnosis, and clinical decision support, and holds significant implications for public health.

% \begin{table}[t]%% placement specifier
% %% Use tabular environment to tag the tabular data.
% %% https://en.wikibooks.org/wiki/LaTeX/Tables#The_tabular_environment
% \centering%% For centre alignment of tabular.
% \begin{tabular}{l c r}%% Table column specifiers
% %% Tabular cells are separated by &
%   1 & 2 & 3 \\ %% A tabular row ends with \\
%   4 & 5 & 6 \\
%   7 & 8 & 9 \\
% \end{tabular}
% %% Use \caption command for table caption and label.
% \caption{Table Caption}\label{fig1}
% \end{table}


% %% For authoryear reference style

% \bibitem[Lamport(1994)]{lamport94}
%   \textit{\LaTeX: a document preparation system},
%   Addison Wesley, Massachusetts,

% \end{thebibliography}
\bibliographystyle{elsarticle-num}
\bibliography{ref}
\end{document}